\documentclass[11pt,journal]{IEEEtran}		% Journal

\IEEEoverridecommandlockouts
\usepackage{cite}
\usepackage{amsmath,amssymb,amsfonts,bm,ntheorem}
\usepackage{mathtools}

\usepackage{algorithm,algorithmic}
\usepackage{graphicx}
\usepackage{textcomp}
\usepackage{xcolor}
\usepackage[small,bf]{caption}
\usepackage{color}
\usepackage{multicol}
\usepackage{rotating}
\usepackage{subfigure}
\usepackage{tablefootnote}
\usepackage[flushleft]{threeparttable}
\usepackage{rotating}
\usepackage{listings}
\usepackage{lipsum}
\usepackage{cuted}
\usepackage{flushend}

\usepackage{array}
\usepackage{framed}
\usepackage{fancybox}
\definecolor{shadecolor}{rgb}{1, 1, 0.5}
\usepackage[shortlabels]{enumitem}

\usepackage{venndiagram}
\usepackage{tikz}
\usetikzlibrary{shapes.geometric, fit,positioning, arrows}

\def\BibTeX{{\rm B\kern-.05em{\sc i\kern-.025em b}\kern-.08em
    T\kern-.1667em\lower.7ex\hbox{E}\kern-.125emX}}
    
\usepackage[bookmarks=false]{hyperref}
\hypersetup{
    colorlinks=false,	% make the links colored
    linkcolor=blue,	% color TOC links in blue
    filecolor=magenta,      
    urlcolor=cyan,	% color URLs in red
    pdftitle={Sharelatex Example},
    bookmarks=true,
    pdfpagemode=FullScreen,
}    

\usepackage{blindtext}

\newcommand{\argmin}{\mathop{\rm argmin}}

\newcommand{\ie}{{\it i.e.}}

\graphicspath{ {./images/} }

\theorembodyfont{\itshape}

\newtheorem{theorem}{Theorem}
\newtheorem{proposition}{Proposition}
\newtheorem{corollary}{Corollary}
\newtheorem{lemma}{Lemma}
\newtheorem{definition}{Definition}
\newtheorem{remark}{Remark}

\begin{document}

%%%%%%%%%%%
% Paper Title
%%%%%%%%%%%
\title{Adaptive Iterative Soft-Thresholding Algorithm \\ with the Median Absolute Deviation
}
%%%%%%%%% role of gamma %%%%%%%%%%%%%%%%%

%%%%%%%%%%%
% Author Names
%%%%%%%%%%%
\author{Yining~Feng,~\IEEEmembership{Member,~IEEE,} and~Ivan~Selesnick,~\IEEEmembership{Fellow,~IEEE}
% \author{ Abdullah~Al-Shabili
% %\thanks{Manuscript received January 20, 2002; revise d August 26, 2015. This work was supported by the IE EE.}
}

\maketitle

% %%%%%%%%%%%%%%%%%%%%%%%%%%%%%%%%%%%%%%%%%%%%%%%%%%
% %%%%%%%%%%%%%%%%%%%%%%%%%%%%%%%%%%%%%%%%%%%%%%%%%%
% %%%%%%%%%%%%%%%%%%%%%%%%%%%%%%%%%%%%%%%%%%%%%%%%%%
% %%%%%%%%%%%%%%%%%%%%%%%%%%%%%%%%%%%%%%%%%%%%%%%%%%
% %%%%%%%%%%%%%%%%%%%%%%%%%%%%%%%%%%%%%%%%%%%%%%%%%%
\begin{abstract}
The adaptive Iterative Soft-Thresholding Algorithm (ISTA) has been a popular algorithm for finding a desirable solution to the LASSO problem without explicitly tuning the regularization parameter $\lambda$.
Despite that the adaptive ISTA is a successful practical algorithm, few theoretical results exist. 
In this paper, we present the theoretical analysis on the adaptive ISTA with the thresholding strategy of estimating noise level by median absolute deviation.
We show properties of the fixed points of the algorithm, including scale equivariance, non-uniqueness, and local stability, prove the local linear convergence guarantee, and show its global convergence behavior.
\end{abstract}

% \begin{IEEEkeywords}

% \end{IEEEkeywords}

%%%%%%%%%%%%%%%%%%%%%%%%%%%%%%%%%%%%%%%%%%%%%%%%%%
%%%%%%%%%%%%%%%%%%%%%%%%%%%%%%%%%%%%%%%%%%%%%%%%%%
%%%%%%%%%%%%%%%%%%%%%%%%%%%%%%%%%%%%%%%%%%%%%%%%%%
%%%%%%%%%%%%%%%%%%%%%%%%%%%%%%%%%%%%%%%%%%%%%%%%%%
%%%%%%%%%%%%%%%%%%%%%%%%%%%%%%%%%%%%%%%%%%%%%%%%%%

\section{Introduction}

Finding a sparse approximate solution of an underdetermined system of equations, $y = Ax$, has been a problem of interest in signal processing, data analysis, and statistics. 
The goal is to estimate the underlying sparse signal $x_0$ from the noisy measurement $y$ of the linear system
\begin{align}
    y = Ax_0 + w, \quad w \sim \mathcal{N}(0, \sigma^2 I),
\end{align}
where $A \in \mathbb{R}^{M \times N}$ is the measurement matrix capturing the forward measurement process, and $w$ is the additive white Gaussian noise.
The estimation $\bar{x}$ can be obtained by solving the LASSO problem \cite{tibshirani1996regression}, a $\ell_1$ norm regularized least square problem
\begin{align} \label{lasso_problem}
    \bar{x} \!=\! \arg \min_{x \in \mathbb{R}^N} \{ (1/2) \|y\! -\! Ax\|_2^2 + \lambda \|x\|_1\}, \, \lambda \in \mathbb{R}^+.
\end{align} 
The celebrated Iterative Soft-Thresholding Algorithm (ISTA) \cite{daubechies2004iterative}, an instance of the forward-backward splitting (FBS) algorithm \cite{combettes2011proximal}, is the first-order algorithm for solving the large-scale LASSO problems  
\begin{align} \label{ISTA}
    x^{k+1} = \text{soft}(x^k - \mu A^T(Ax^k - y ), \mu \lambda ), \, 0 \leqslant \mu \leqslant 2/\|A\|_2^2,
\end{align}
where $\text{soft}(x, \lambda) = \text{sign}(x)(|x| - \lambda)_+$ is the element-wise soft-thresholding operator, and $\mu$ is the step size. 
Many sparse approximation problems in machine learning and signal processing can be obtained as the solution to the LASSO problem, which can be solved by ISTA. 

Despite its popularity, tuning $\lambda$ to obtain a solution with desired properties while not compromising computational cost still poses an obstacle in practice and remains an open problem. From the perspective of the objective function, the least-angle regression (LARS) algorithm \cite{efron2004least,tibshirani2013lasso} computes the LASSO solution path for all possible $\lambda$s and selects the solution that minimizes the Stein's unbiased risk estimate (SURE) of the objective function in \eqref{lasso_problem} \cite{tibshirani2012degrees}. The obtained LASSO solution is optimal in the mean-squared-error (MSE) sense with minimum assumptions, but LARS is not competitive in terms of computation time for large-scale problems \cite{maleki2010optimally}. 

On the algorithm side, variations of ISTA were proposed to find a LASSO solution (under the assumption that the algorithms will converge),
\begin{align} \label{adaptive_ISTA}
    x^{k+1} = \text{soft}(x^k - \mu A^T(Ax^k - y ), \lambda^k ), \quad \mu \in \mathbb{R}^+,
\end{align}
with different adaptive thresholding strategies for choosing the thresholds $\lambda^k$. The adaptive thresholds $\lambda^k$ are not chosen based on regularization parameter $\lambda$ in \eqref{lasso_problem}, but instead selected according to $x^k - \mu A^T(Ax^k - y )$ in each iteration. There are two successful heuristics: 1) heuristic of $K$-sparse \cite{blumensath2009iterative,needell2009cosamp,dai2009subspace} assumes prior knowledge of the true sparsity of the underlying signal and sets the thresholds to the magnitude of the $(K+1)^{th}$ largest element; 2) heuristic of multiple access interference noise \cite{donoho2012sparse} models $x^k - \mu A^T(Ax^k - y )$ as sparse signal plus additive noise, estimates the noise level $\sigma^k$ by the standard deviation and sets the thresholds accordingly. 

\begin{table*}[t] 
\centering
\resizebox{2\columnwidth}{!}{
\begin{tabular}{ | c | c| c  | } 
  \hline
  & $K$-sparse & Noise level $\sigma$ by standard deviation \\ 
  \hline
  Random  & exact recovery, local convergence \cite{maleki2009coherence}   & LASSO solution correspondence \cite{donoho2011noise}, \\
  design $A$ & global convergence \cite{wang2015linear} & global convergence (infinite dimension) \cite{bayati2011lasso} \\
  \hline
  Deterministic  & restricted optimal,  & local stability  \\ 
  design $A$ & negative convergence guarantee \cite{barber2018gradient,liu2020between} & (finite dimension)\cite{ma2021local,rangan2019convergence} \\
  \hline
\end{tabular}
}
\caption{Theoretical analysis of adaptive ISTA with different settings and thresholding strategies}
\label{lit_review}
\end{table*}

The adaptive ISTA is widely adopted in practice due to its simplicity. 
As similar as it seems to ISTA \eqref{ISTA}, adaptive ISTA is not an optimization algorithm solving for a minimizer of an objective function, but a fixed point iteration. The existence of its fixed point, the properties of the fixed points\footnote{That is, what kind of LASSO solution the algorithm converges to and how many fixed points exist.} and the convergence behavior have to be analyzed on a case by case base with various assumptions and techniques.
And few theoretical analyses exist, especially for the deterministic design $A$ (or structured measurement matrix).
From the compressive sensing literatures focusing on the random design $A$ with $K$-sparse strategy, under the exact recovery assumption\footnote{Exact recovery assumes that the LASSO solution $\bar{x}$ with appropriate $\lambda$ recovers the ground truth $x_0$.}, coherence-based analyses \cite{maleki2009coherence,wang2015linear} establish local and global convergence. However, exact recovery is a rather restrictive assumption that only holds for noiseless and low noise-level scenarios. With the help of the relative concavity notion, the analyses in \cite{barber2018gradient,liu2020between} give a negative convergence result for deterministic design with $K$-sparse strategy. This means that, in the worse case, the adaptive ISTA does not converge to the optimal restricted to a pre-defined sparsity level.   

Most of the theoretical analyses for multiple access interference noise strategy come from works on Approximate Message Passing\footnote{AMP is technically a variant of the Douglas-Rachford Splitting Algorithm (or the alternating direction method of multipliers, ADMM), with adaptive thresholding strategy. We refer to it here because ISTA with such a strategy has terrible convergence behavior.}(AMP). Specifically, a one-to-one correspondence between the fixed point of AMP and LASSO solution is established in \cite{donoho2011noise} for the random design $A$, and a global convergence guarantee is established for the infinite dimension case in \cite{bayati2011lasso}. For the deterministic design, only the stability analyses of the fixed point exist \cite{ma2021local,rangan2019convergence}. Table \ref{lit_review} summarizes the theoretical results.

In this paper, we investigate the adaptive ISTA with the thresholding strategy of estimating the noise level with the Median Absolute Deviation (MAD),
\begin{align} \label{AISTA} \nonumber
    x^{k+1} & = T(x^k - \mu A^T(Ax^k - y )), \quad 0 \leqslant \mu \leqslant 2/\|A\|_2^2, \\
    & T(z)  = \text{soft}(z, \gamma \text{median}(|z|)), \quad \gamma > 1. 
\end{align}
The MAD $\text{median}(|\cdot|)$ was first introduced to signal processing in \cite{donoho1995adapting}, as a robust estimator of the noise level $\sigma$. It proved to be more resilient to sparse outliers than the standard deviation. The noise level of an independent and identically distributed (i.i.d.) Gaussian random vector $w$ can be estimated as
\begin{align}
\sigma \!=\! \text{median}(|w|)/\Phi^{-1}(3/4) \!\approx\! 1.4826 \, \text{median}(|w|),
\end{align}
where $\Phi^{-1}$ is the inverse of the cumulative distribution function for the standard normal distribution.
It is subsequently used for the wavelet transform denoising \cite{luisier2007new,beheshti2010noise} and AMP \cite{donoho2009message,maleki2010analysis} and has a deep connection to the SURE estimator. Empirically, the ISTA with the MAD thresholding strategy has stable convergence behavior and performs well for both random and deterministic design matrices under various noise levels. 
However, the theoretical understanding of its behaviors is largely limited.

In this paper, we make the following contributions on theoretical and empirical aspects:
\begin{enumerate}
\item We establish the correspondence between the fixed points of \eqref{AISTA} and the LASSO solutions through the fixed point condition and show properties of the fixed points, including scale equivalence, non-uniqueness, and local stability. 
\item We prove the local linear convergence guarantee of \eqref{AISTA} and show its global convergence behavior. 
\item We show the performance and summarize its behavior for finite-dimension yet large-scale numerical experiments on i) the compressive sensing problem, ii) the linear inverse problem with an undersampled DCT matrix, and iii) the sparse deconvolution problem. 
\end{enumerate}
Our assumptions throughout this paper are minimal. We place no assumptions on the measurement matrices $A$, considering them fixed and nonrandom.

\section{Property of the Fixed Points}

First, we have to reemphasize, even though iteration \eqref{AISTA} is quite similar to \eqref{ISTA}, it is not a numerical algorithm for solving an optimization problem, because there is no objective function to be minimized. Instead, iteration \eqref{AISTA} is a fixed point iteration selecting a solution from all possible LASSO solutions, as will be further discussed in this section in the characterization of the properties of the fixed points.    

\subsection{Fixed Point Condition}

In this subsection, we introduce a theorem describing the fixed point condition. All the fixed points of iteration \eqref{AISTA} have to satisfy the fixed point condition. 

First, we define the subgradient of the $\ell_1$ norm.
\begin{definition} \label{subgradient_L1}
The subgradient $\partial \|x\|_1 \in \mathbb{R}^N$ of $\ell_1$ norm $\|x\|_1$ is a set-valued function defined element-wise as
\begin{align}
    \partial \|x_i\|_1 = \begin{cases}
        1 & x_i > 0, \\
        (-1, \, 1) & x_i = 0, \\
        -1 & x_i < 0.
    \end{cases} 
\end{align}
\end{definition}

We also note that the soft-thresholding operator with the MAD has the following properties.
\begin{lemma} \label{operator_T}
    The soft-thresholding operator with the MAD $T(z)  = \text{\normalfont soft}(z, \gamma \text{\normalfont median}(|z|))$ is the solution to the minimization problem
    \begin{align} \label{MAD_thresholding}
        \arg \min_x \left\{ (1/2)\|x - z\|_2^2 + \gamma\text{\normalfont median}(|z|) \|x\|_1  \right\}.
    \end{align}
    Given $\gamma > 1$, the size of the support of $T(z)$ is smaller than $N/2$. 
\end{lemma}
\begin{IEEEproof}
Take the derivative of the objective function in \eqref{MAD_thresholding}, we obtain the optimality condition
\begin{align*}
     0 &\in x - z + \gamma\text{median}(|z|) \,  \partial \|x\|_1 \\
    \implies x &\in z - \gamma\text{median}(|z|) \,  \partial \|x\|_1,
\end{align*}
and it can be written as 
\begin{align*}
    x_i = \begin{cases}
        z_i - \gamma\text{median}(|z|) & z_i > \gamma\text{median}(|z|), \\
        0 & |z_i| < \gamma\text{median}(|z|), \\
        z_i + \gamma\text{median}(|z|) & z_i < -\gamma\text{median}(|z|),
    \end{cases} 
\end{align*}
the explicit formula of $T(z)$. For $\gamma > 1$, there are more than half of the coordinates are set to 0, since $|z_i| < \gamma\text{median}(|z|)$, thus the number of non-zeros elements is smaller than $N/2$. 
\end{IEEEproof}

\begin{theorem} \label{theorem_fixed_point}
Given $\gamma > 1$, any fixed point $x^*$ of iteration \eqref{AISTA} satisfying the fixed point condition
\begin{align} \label{fixed_point_condition}
- A^T(A x^* - y) \in \gamma \text{\normalfont median}(|-A^T(A x^* - y)|) \, \partial \|x^*\|_1,
\end{align}
is a LASSO solution, and is invariant to $\mu$. 
\end{theorem}
\begin{IEEEproof}
Assume that the iteration has a fixed point $x^*$, it is the solution to the following problem according to Lemma \ref{operator_T}
\begin{align*}
    x^* = \argmin_z \{ &(1/2) \|z - (x^* - \mu A^T(Ax^* - y ) )\|_2^2 \\
    &+ \gamma \text{median}(|x^* - \mu A^T(Ax^* - y )|) \|z\|_1 \},
\end{align*}
% which is completely different from how we derive the proximal gradient descent. 
let $z = x^*, x = x^* - \mu A^T(Ax^* - y )$, the corresponding optimality condition is
\begin{align*} 
   0  &\in  x^*  - (x^* - \mu A^T(Ax^* - y )) \\
   & \quad \quad + \gamma \text{median}(|x^* - \mu A^T(Ax^* - y )|) \partial \|x^*\|_1, \\
   0  &\in \mu A^T(Ax^* - y ) \\
   & \quad \quad + \gamma \text{median}(|x^* - \mu A^T(Ax^* - y )|) \partial \|x^*\|_1. 
\end{align*}
By Definition \ref{subgradient_L1}, coordinate-wise,
\begin{align} \label{3cases}
       &-\mu [A^T(A x^* - y)]_i \\ \nonumber
       =&\begin{cases}
       \gamma \text{median}(|x^* - \mu A^T(A x^* - y)|), & \text{if} \ x^*_i > 0, \\
    \gamma \text{median}(|x^* - \mu A^T(A x^* - y)|) \partial \|x^*_i\|_1, &\text{if} \ x^*_i = 0, \\
     -\gamma \text{median}(|x^* - \mu A^T(A x^* - y)|), & \text{if} \ x^*_i < 0,
      \end{cases}
\end{align}
thus we have
\begin{align*}
    \text{sign}(-\mu [A^T(A x^* - y)]_i) =  \text{sign}(x^*_i) = \text{sign}(\partial \|x^*_i\|_1).
\end{align*}
As a result
\begin{align*}
       |[x^* - \mu A^T(A x^* - y)]_i| & = |x_i^* - \mu [A^T(A x^* - y)]_i| \\
       &= |x_i^*| + |- \mu [A^T(A x^* - y)]_i| \, \forall i.
\end{align*}
We can further have the following simplification combining \eqref{3cases}
\begin{align*}
    &|x_i^*| + |- \mu [A^T(A x^* - y)]_i| \\
    = &\begin{cases}
        |x_i^*| + \gamma \text{median}(|x^* - \mu A^T(A x^* - y)|) & \text{if} \ |x^*_i| \not = 0, \\
        \gamma \text{median}(|x^* - \mu A^T(A x^* - y)|) \left|\partial \|x^*_i\|_1 \right| & \text{if} \ |x^*_i| = 0.
    \end{cases}
\end{align*}
Note that since $\left|\partial \|x^*_i\|_1 \right| < 1$, the magnitudes of the coordinates from the first case are strictly greater than those from the second case. 
That is, the magnitudes of $|x_i^*| + |- \mu [A^T(A x^* - y)]_i|$ for on-support coordinates ($|x^*_i| \not = 0$) are strictly greater than the off-support ($|x^*_i| = 0$) ones.

% Also note that $\forall x^*_i \not = 0$, $|-\mu [A^T(A x^* - y)]_i|$ admits the same absolute value, \ie \, maximum value of the vector $|-~\mu A^T(A x^* - ~y)|$; and $\forall x^*_i = 0$, $|-\mu [A^T(A x^* - y)]_i|$ are strictly smaller than the maximum value. Let $|\theta|_{(1)} \le \cdots \le |\theta|_{(N)}$ be the order statistics of $|\theta_1|, \cdots, |\theta_N|$ of a vector, 
% \begin{align*}
%     |[x^* - \mu A^T(A x^* - y)]_{(i)}| =  |x_{(i)}^*| + |- \mu [A^T(A x^* - y)]_{(i)}|, \quad \forall x^*_{(i)} \not = 0, \\
%     |[x^* - \mu A^T(A x^* - y)]_{(i)}| = |- \mu [A^T(A x^* - y)]_{(i)}|, \quad \forall x^*_{(i)} = 0
% \end{align*}
By Lemma \ref{operator_T}, because $\gamma > 1$, the support of $x^*$ is always less than $N/2$, then the median index $j$ always occurs at an off-support coordinate, 
\begin{align*}
    & |x_j^*| + |- \mu [A^T(A x^* - y)]_j| \\
    = &\gamma \text{median}(|x^*-\mu A^T(A x^*-y)|) \left|\partial \|x^*_j\|_1 \right|  \\
    = & | -\mu [A^T(A x^* - y)]_j| \\
  \implies \quad & \text{median}(|x^*| + |- \mu A^T(A x^* - y)|) \\
   & \quad \quad \quad \quad \quad \quad \quad \quad= \text{median}(|- \mu A^T(A x^* - y)|),
\end{align*}
therefore
\begin{align*} 
       & -\mu A^T(A x^* - y) \in \gamma \text{median}(|- \mu A^T(A x^* - y)|) \partial\|x^*\|_1  \\
       \implies & - A^T(A x^* - y) \in \gamma \text{median}(|- A^T(A x^* - y)|) \partial\|x^*\|_1.
\end{align*}
That is, the fixed point is only dependent on the choice of $\gamma$, and invariant to $\mu$. 

Let 
\begin{align*}
    \lambda^* = \gamma \text{median}(|- A^T(A x^* - y)|),
\end{align*}
the fixed point condition is 
\begin{align*}
    - A^T(Ax^* - y ) \in \lambda^*\partial \|x^*\|_1,
\end{align*}
which is the optimality condition of a LASSO problem. It indicates that $x^*$ is a LASSO solution corresponding to certain $\lambda^*$.
\end{IEEEproof}

\begin{remark} \label{remark_lasso_correspond}
Let the set $\mathcal{S}$ 
\begin{align} \nonumber
     \mathcal{S} := &\bigg\{\bar{x}(\lambda))\bigg| \bar{x}(\lambda) = \argmin_z \{(1/2)\|y - Az\|_2^2 \\
     &\quad \quad \quad \quad \quad \quad \quad \quad \quad  + \lambda \|z\|_1 \},  \, \forall \lambda \ge 0 \bigg\}
\end{align}
be the set containing all the solutions to the LASSO problem with $\lambda > 0$, then the fixed point condition \eqref{fixed_point_condition} selects $x^*$ from the LASSO solution set $\mathcal{S}$. 
%Since $\bar{x} \in \mathcal{S}$ is not necessarily unique with respect to $\lambda$, the fixed point $x^*$ is not unique given $\gamma$. 
\end{remark}

\begin{corollary}
Given $\gamma$, the fixed point $x^*$ is scale equivariance to the observation $y$,
\begin{align}
x^*(c y) = c x^*(y), \quad c > 0.
\end{align}
\end{corollary}

\begin{IEEEproof}
Let $\hat{y} = c y, c > 0$ be the scaled version of the observation $y$, $x^*$ be the fixed point with respect to $y$. We have  
\begin{align*}
& \!-\! A^T(A x^* \!-\!  y)\! \in\! \gamma \text{median}(|\!-\! A^T(A x^* \!-\! y)|) \partial\| x^*\|_1 \\
\implies & \!-\! A^T(A c x^* \!-\! c y) \\
& \quad \quad \quad \quad \quad \! \in\! \gamma \text{median}(|\!-\! A^T(A c x^* \!-\! c y)|) \partial\|c x^*\|_1 
\end{align*}
Let $\hat{x}^* = c x^*$, we obtain the same fixed point condition in \eqref{fixed_point_condition},
\begin{align*}
& - A^T(A \hat{x}^* - \hat{y}) \in \gamma \text{median}(|- A^T(A \hat{x}^* - \hat{y})|) \partial\|\hat{x}^*\|_1.
\end{align*}
\end{IEEEproof}

% cite Boosting the Performance of Plug-and-Play Priors via Denoiser Scaling

Scaling the observation $y$ simultaneously scales the signal strength and noise level, and thus does not change the signal-to-noise ratio (SNR).
This corollary suggests that we can scale the proposed estimator from one noise level to another given the same SNR. 
This scale equivariance property also holds for optimal estimators such as MMSE estimator \cite{xu2020boosting}, and LASSO solution found by Tunable Approximate Message Passing (AMPT)\cite{donoho2011noise}.

\subsection{Non-uniqueness of the Fixed Points}

\begin{figure}[t!]
    \centering
    \includegraphics[width=0.45\textwidth]{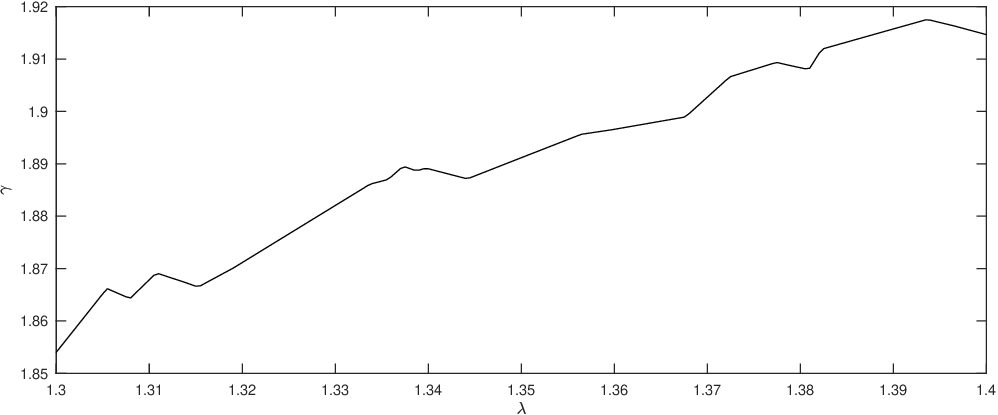}
    \caption{$\gamma$ is a continuous piecewise rational function with respect to $\lambda$ \label{fig_gamma_lambda}}
\end{figure}

As suggested by Theorem \ref{theorem_fixed_point} and Remark \ref{remark_lasso_correspond}, all LASSO solutions are candidate fixed points of the iteration \eqref{AISTA}, and any fixed point always corresponds to a LASSO solution. However, the correspondence is not unique, that is, the fixed point regarding one $\gamma$ can correspond to LASSO solutions regarding multiple $\lambda$s. As a result, the fixed point is not unique. The correspondence is specified by $\lambda^* = \gamma \text{median}(|- A^T(A x^* - y)|)$, or
\begin{align} \label{gamma_lambda}
    \gamma = \frac{\lambda}{\text{median}(|- A^T(A \bar{x}(\lambda) - y)|)},
\end{align}
where $\bar{x}(\lambda)$ is the LASSO solution given $\lambda$. We show the non-unique correspondence by investigating $\gamma$ as a function of $\lambda$. 
For each $\lambda$, there exists a LASSO solution $\bar{x}$, and $\text{median}(|- A^T(A \bar{x}(\lambda) - y)|)$, thus $\gamma$ is a function of $\lambda$. 
Fig.~\ref{fig_gamma_lambda} shows the function $\gamma(\lambda)$.
We characterize the correspondence as follows. 

We recall the definition of the equicorrelation set and equicorrelation sign of the LASSO solution \cite{tibshirani2013lasso}.
\begin{definition}
The unique equicorrelation set $\mathcal{I}$ of the LASSO solution at one $\lambda$ is defined as
\begin{align}
\mathcal{I} = \{ i \in \{1, \cdots, N\}: |-A_i^T (A \bar{x} -y ) | = \lambda\},
\end{align}
and the equicorrelation signs $s$ by
\begin{align}
 s = \text{sign}(-A_i^T (A \bar{x} -y )). 
\end{align}
\end{definition}

\begin{lemma}
Given $y, A$, and let $\bar{x}(\lambda)$ be the LASSO solution regarding to $\lambda$, $\text{median}(|- A^T(A \bar{x}(\lambda) - y)|)$ is a continuous piecewise linear function with respect to $\lambda$, and can be written as
\begin{align} \label{median}
    \text{\normalfont median}(|- A^T(A \bar{x}(\lambda) - y)|) = |a \lambda + b|, \, \lambda \in [\lambda_k, \lambda_{k+1}),
\end{align}
where 
\begin{align} \label{variable_a}
a & =  A_j^T A_{\mathcal{I}} (A_{\mathcal{I}}^T A_{\mathcal{I}})^{-1}s, \\ \label{variable_b}
b & = A_j^T (I - A_{\mathcal{I}} (A_{\mathcal{I}}^T A_{\mathcal{I}})^{-1}A_{\mathcal{I}}^T )y.
\end{align}
are constants when the median index $j$ and equicorrelation set $\mathcal{I}$ does not change within the interval $\lambda_k, \lambda_{k+1}$.
\end{lemma}

\begin{IEEEproof}
First, we note that the optimal subgradient of a LASSO solution $- A^T(A \bar{x}(\lambda) - y): \mathbb{R}^N \to \mathbb{R}^N$ is a unique continuous piecewise linear mapping regarding to $\lambda$ \cite{tibshirani2013lasso}, so is $\text{median}(|- A^T(A \bar{x}(\lambda) - y)|)$ from composition of continuous piecewise linear functions. % reference?

For any LASSO solution, all coordinates outside the equicorrelation set $\mathcal{I}$ is zero, \ie \, $\bar{x}_{-\mathcal{I}} = 0$, and thus the optimality condition is equivalent to 
\begin{align*}
- A_{\mathcal{I}}^T (A_{\mathcal{I}} \bar{x}_{\mathcal{I}} - y) = \lambda s.
\end{align*}
Solve for $\bar{x}_{\mathcal{I}}$, we get
\begin{align*}
\bar{x}_{\mathcal{I}} = (A_{\mathcal{I}}^T A_{\mathcal{I}})^{-1}(A_{\mathcal{I}}^T y - \lambda s).
\end{align*}
Suppose $\text{median}(|- A^T(A \bar{x}(\lambda) - y)|)$ occurs at index $j$, then 
\begin{align*}
&\text{median}(|- A^T(A \bar{x}(\lambda) - y)|) \\
= &|-A_j^T( A_{\mathcal{I}} \bar{x}_{\mathcal{I}} - y)| \\
= &| A_j^T A_{\mathcal{I}} (A_{\mathcal{I}}^T A_{\mathcal{I}})^{-1} s \lambda  + A_j^T (I - A_{\mathcal{I}} (A_{\mathcal{I}}^T A_{\mathcal{I}})^{-1}A_{\mathcal{I}}^T )y |.
\end{align*}
When $A_j, A_{\mathcal{I}}$ does not change, the terms $A_j^T A_{\mathcal{I}} (A_{\mathcal{I}}^T A_{\mathcal{I}})^{-1} s$ and $A_j^T (I - A_{\mathcal{I}} (A_{\mathcal{I}}^T A_{\mathcal{I}})^{-1}A_{\mathcal{I}}^T )y$ are constants.
Thus, $\text{median}(|- A^T(A \bar{x}(\lambda) - y)|)$ is linear with respect to $\lambda$ whenever the median index $j$ and equicorrelation set $\mathcal{I}$ does not change, and thus piecewise linear. 
\end{IEEEproof}

\begin{figure}[t!]
    \centering
    \includegraphics[width=0.5\textwidth]{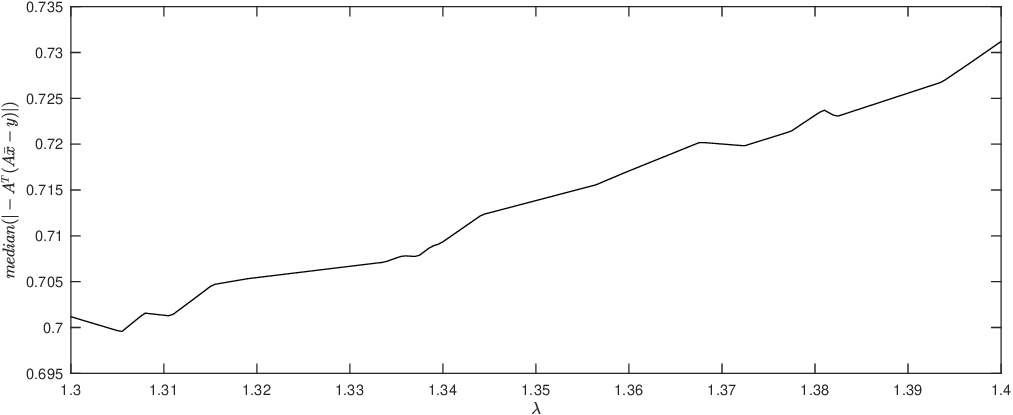}
    \caption{$\text{median}(|- A^T(A \bar{x}(\lambda) - y)|)$ is a continuous piecewise linear function with respect to $\lambda$ \label{fig_median_lambda}}
\end{figure}

% Let $f(\lambda) = |a \lambda + b|, \lambda \in [\lambda_k, \lambda_{k+1})$ be the linear function captures the linear relationship between $\text{median}(|- A^T(A \bar{x}(\lambda) - y)|)$ and $\lambda$, where the median index $j$ and equicorrelation set $\mathcal{I}$ does not change in between $\lambda_k, \lambda_{k+1}$. We have 
% \begin{align}
% a & =  A_j^T A_{\mathcal{I}} (A_{\mathcal{I}}^T A_{\mathcal{I}})^{-1}s, \\
% b & = A_j^T (I - A_{\mathcal{I}} (A_{\mathcal{I}}^T A_{\mathcal{I}})^{-1}A_{\mathcal{I}}^T )y.
% \end{align}

% Recall that $y = Ax_0 + w$, and decompose $Ax_0 = A_{\mathcal{I}} [x_0]_{\mathcal{I}} + A_{-\mathcal{I}} [x_0]_{-\mathcal{I}} $, then
% \begin{align}
% b & = A_j^T (I - A_{\mathcal{I}} (A_{\mathcal{I}}^T A_{\mathcal{I}})^{-1}A_{\mathcal{I}}^T )Ax_0 +  A_j^T (I - A_{\mathcal{I}} (A_{\mathcal{I}}^T A_{\mathcal{I}})^{-1}A_{\mathcal{I}}^T )w \\
% & = A_j^T (I - A_{\mathcal{I}} (A_{\mathcal{I}}^T A_{\mathcal{I}})^{-1}A_{\mathcal{I}}^T )(A_{\mathcal{I}} [x_0]_{\mathcal{I}} + A_{-\mathcal{I}} [x_0]_{-\mathcal{I}}) +  A_j^T (I - A_{\mathcal{I}} (A_{\mathcal{I}}^T A_{\mathcal{I}})^{-1}A_{\mathcal{I}}^T )w \\
% & = A_j^T (I - A_{\mathcal{I}} (A_{\mathcal{I}}^T A_{\mathcal{I}})^{-1}A_{\mathcal{I}}^T ) A_{-\mathcal{I}} [x_0]_{-\mathcal{I}} +  A_j^T (I - A_{\mathcal{I}} (A_{\mathcal{I}}^T A_{\mathcal{I}})^{-1}A_{\mathcal{I}}^T )w.
% \end{align}

Fig.~\ref{fig_median_lambda} illustrates the continuous piecewise linear behavior.

\begin{proposition} \label{multiple_fixed_points}
Let $\gamma$ be a function of $\lambda$ defined by \eqref{gamma_lambda}, then $\gamma$ is a continuous piecewise rational function, and is not monotone. The correspondence between $\gamma$ and $\lambda$ is not unique, and thus the fixed point $x^*$ is not unique. 
\end{proposition}

\begin{IEEEproof}
Consider the function $\gamma(\lambda) = \frac{\lambda}{ |a \lambda + b |}$, where $|a\lambda + b|$ captures the linear relationship between $\text{median}(|- A^T(A \bar{x}(\lambda) - y)|)$ and $\lambda$. Then $\gamma'(\lambda) = \frac{b \text{sign}(a \lambda + b)}{ (a \lambda + b)^2 }$ can be positive or negative, depending on the sign of $b(a \lambda + b)$. That is, the function $\gamma(\lambda)$ can be increasing or decreasing in each interval $[\lambda_k, \lambda_{k+1})$. Because $\gamma(\lambda)$ is not monotone, there exist multiple $\lambda$s given certain $\gamma$. Hence, we conclude that the fixed point $x^*$ is not unique.
\end{IEEEproof}

% Fig. 2 illustrates the relationship between $\gamma$ and $\lambda$. 

\begin{figure}[t!]
    \centering
    \includegraphics[width=0.23\textwidth]{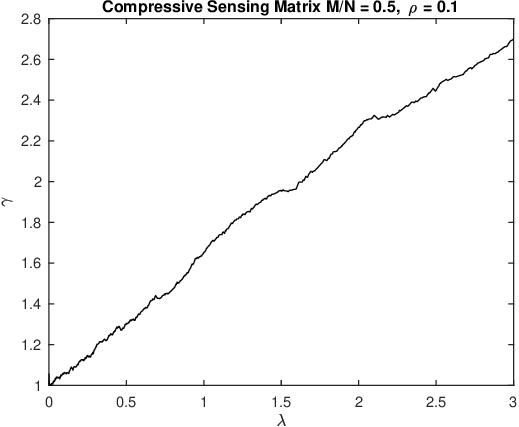}
    \includegraphics[width=0.23\textwidth]{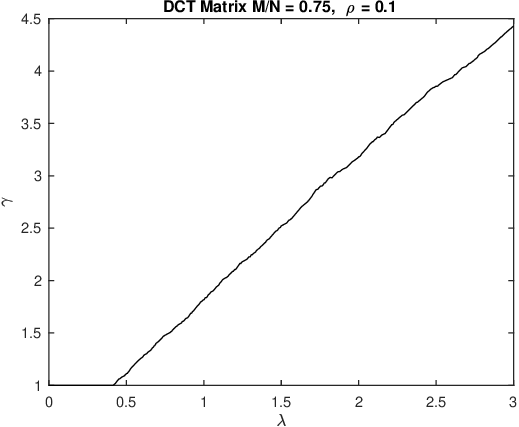}
    \includegraphics[width=0.23\textwidth]{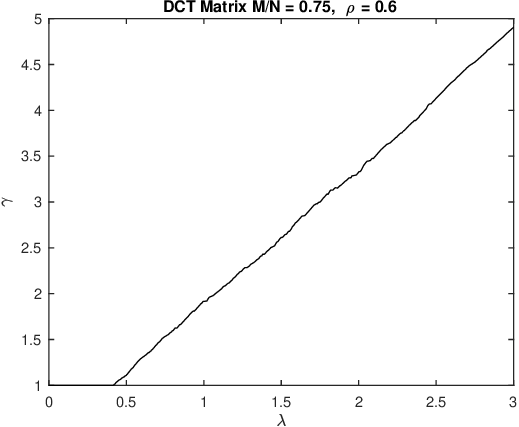}
    \includegraphics[width=0.23\textwidth]{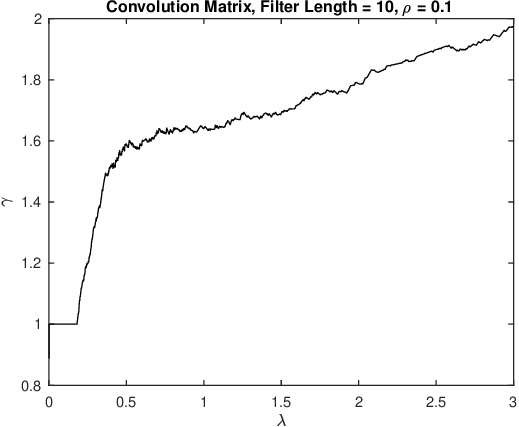}
    \caption{$\gamma(\lambda)$ for big range of $\lambda$s in different settings (design matrices and sparsity levels). \label{gam_lam_correspondence}}
\end{figure}

\begin{remark} \label{one_to_one}
The non-uniqueness of the fixed points might not be ideal, because we do not know which one fixed point the algorithm would converge to. But in practice, it is not a huge issue, because the non-uniqueness only occurs within a small range of $\lambda$s, that is, the distance between two $\lambda$s corresponding to one $\gamma$ is small. As a result, the multiple fixed points would not be so different from each other. In a big range of $\lambda$s, the $\gamma$ and $\lambda$ correspondence is approximately one-to-one, we show the correspondence in several different scenarios in Fig.~\ref{gam_lam_correspondence}.  
\end{remark}

\subsection{Stability of the Fixed Points}

From the discussion in the last subsection, there exist multiple LASSO solutions satisfying the fixed point condition \eqref{fixed_point_condition} as the candidate fixed points. However, not all of the candidate LASSO solutions are stable fixed points of the proposed algorithm. We will further characterize the stable fixed points in the following proposition. 

We first introduce the following lemma on nonsymmetric rank-1 update of a positive definite matrix.
\begin{lemma} \label{nonsymmetric_positve_definite}
    Given $C \succ 0$, the nonsymmetric matrix $C + uv^T$ is positive definite only if $1 + v^T C^{-1} u > 0$.
\end{lemma}
\begin{IEEEproof}
    If the nonsymmetric matrix $C + uv^T$ is positive definite, all its eigenvalues are positive. Because the determinant of a matrix is equal to the product of its eigenvalues, the determinant of $C + uv^T$ has to be positive. According to (6.2.3) in \cite{meyer2023matrix}, 
    \begin{align*}
        \det (C + uv^T) = \det(C) (1 + v^T C^{-1} u). 
    \end{align*}
    Because $C \succ 0$, $\det(C) >0$, then $1 + v^T C^{-1} u$ has to be positive.
\end{IEEEproof}

\begin{remark}
    Lemma \ref{nonsymmetric_positve_definite} is a necessary condition, that is, if $C + uv^T \succ 0$ then $1 + v^T C^{-1} u > 0$. Yet it is not a necessary and sufficient condition. We recognize that when $1 + v^T C^{-1} u > 0$, $\det (C + uv^T) >0$ but $C + uv^T \not \succ 0$, which can happen when $C + uv^T$ has even number of eigenvalues with negative real part. The complication is due to the fact that $C + uv^T$ is nonsymmetric.  
\end{remark}

\begin{proposition} \label{prop_stability}
Let $\gamma$ be a function of $\lambda$ defined by \eqref{gamma_lambda}, and there are multiple $\lambda$s and LASSO solutions corresponding to one $\gamma$. The LASSO solutions of $\lambda$s are stable fixed points of the proposed algorithm only if $\gamma(\lambda)$ is increasing (or $\gamma'(\lambda) > 0$). 
\end{proposition}

\begin{IEEEproof}
% Recall the thresholding operator $T: \mathbb{R}^N \to \mathbb{R}^N$ is defined as $T(z) = \text{soft}(z, \gamma \text{median}(|z|))$. Specifically, for $T(z_i) ~= 0, \forall i$, we can write it as 
% \begin{align}
%     T(z_i) = \begin{cases}
%     z_i - \gamma |z_j| & z_i > 0, \\
%     z_i + \gamma |z_j| & z_i < 0,
%     \end{cases}
% \end{align}
% where $\text{median}(|z|) = |z_j|$, or
% \begin{align}
%     T(z) = z - \gamma \, |z_j| \, \text{sign}(z),
% \end{align}
% where $\text{sign}(z)$ is a column vector.
% Then, given $\gamma$, the Jacobian of $T$ at $z$ on the support of $T(z)$ is 
% \begin{align}
%     J_T(z) = I - \gamma \, \text{sign}(z_j) \, \text{sign}(z) \, e_j^T
% \end{align}
% where $e_j$ is a column vector where all but the $i$th coordinates are zeros and the $i$th coordinate is one.
Consider the iteration at its fixed point, 
\begin{align*}
    x^* = T(x^* - \mu A^T(A x^* -y) ).
\end{align*}
Because at the fixed point, the only active coordinates are on its support $x^*_{\mathcal{I}}$, the update can be simplified as
\begin{align*}
    x_{\mathcal{I}}^* &= x_{\mathcal{I}}^* - \mu A_{\mathcal{I}}^T(A_{\mathcal{I}} x_{\mathcal{I}}^* -y) \\
    & \, \,-\! \gamma |[x^*\! -\! \mu A^T(A x^* \!-\!y)]_j| \, \text{sign}(x_{\mathcal{I}}^* \!-\! \mu A_{\mathcal{I}}^T(A_{\mathcal{I}} x_{\mathcal{I}}^* \!-\!y)) \\
    & = x_{\mathcal{I}}^* \! -\! \mu A_{\mathcal{I}}^T(A_{\mathcal{I}} x_{\mathcal{I}}^* \! -\!y) \! -\! \gamma \mu |\! -\!A_j^T(A_{\mathcal{I}} x_{\mathcal{I}}^* \! -\!y)| \, \text{sign}(x_{\mathcal{I}}^* ) \\
    & = x_{\mathcal{I}}^* - \mu A_{\mathcal{I}}^T(A_{\mathcal{I}} x_{\mathcal{I}}^* -y) - \gamma \mu |-A_j^T(A_{\mathcal{I}} x_{\mathcal{I}}^* -y)| \, s
\end{align*}
The Jacobian can be computed 
\begin{align*}
    J(x_{\mathcal{I}}^*) = I - \mu A_{\mathcal{I}}^T A_{\mathcal{I}} + \gamma \mu \text{sign}(-A_j^T(A_{\mathcal{I}} x_{\mathcal{I}}^* -y)) s A^T_j A_{\mathcal{I}}. 
\end{align*}
The local stability of the proposed iteration around its fixed point is determined by the largest eigenvalue (in modulus) of the Jacobian matrix evaluated at $x^*$, that is, the eigenvalues should be within the unit circle of the complex plane, or
\begin{align*}
     I \succ J(x_{\mathcal{I}}^*) \succ -I,
\end{align*}
thus we require the following stable conditions 
\begin{align} \label{stable_con_1}
    & \mu A_{\mathcal{I}}^T A_{\mathcal{I}} - \gamma \mu \text{sign}(-A_j^T(A_{\mathcal{I}} x_{\mathcal{I}}^* -y)) s A^T_j A_{\mathcal{I}} \succ 0, \\ \label{stable_con_2}
    & 2I - \mu A_{\mathcal{I}}^T A_{\mathcal{I}} + \gamma \mu \text{sign}(-A_j^T(A_{\mathcal{I}} x_{\mathcal{I}}^* -y)) s A^T_j A_{\mathcal{I}} \succ 0. 
\end{align}

% \emph{Stable Condition \eqref{stable_con_1}}: Because $A_{\mathcal{I}}^T A_{\mathcal{I}} \succ 0 $ and thus invertible, after similarity transform by $(A_{\mathcal{I}}^T A_{\mathcal{I}})^{\frac{1}{2}}$, we have 
% \begin{align*}
%     & A_{\mathcal{I}}^T A_{\mathcal{I}} - \gamma \text{sign}(-A_j^T(A_{\mathcal{I}} x_{\mathcal{I}}^* -y)) s A^T_j A_{\mathcal{I}} \succ 0 \\
%  \implies & A_{\mathcal{I}}^T A_{\mathcal{I}} - (A_{\mathcal{I}}^T A_{\mathcal{I}})^{\frac{1}{2}}(\gamma \text{sign}(-A_j^T(A_{\mathcal{I}} x_{\mathcal{I}}^* -y)) s A^T_j A_{\mathcal{I}} ) (A_{\mathcal{I}}^T A_{\mathcal{I}})^{-\frac{1}{2}} \succ 0 \\
%  \implies & (A_{\mathcal{I}}^T A_{\mathcal{I}})^{\frac{1}{2}} (I - \gamma \text{sign}(-A_j^T(A_{\mathcal{I}} x_{\mathcal{I}}^* -y)) s A^T_j A_{\mathcal{I}} (A_{\mathcal{I}}^T A_{\mathcal{I}})^{-1} )(A_{\mathcal{I}}^T A_{\mathcal{I}})^{\frac{1}{2}} \succ 0,
% \end{align*}
% and the stable condition \eqref{stable_con_1} is equivalent to 
% \begin{align} \label{stable_con_3}
%     I - \gamma \text{sign}(-A_j^T(A_{\mathcal{I}} x_{\mathcal{I}}^* -y)) s A^T_j A_{\mathcal{I}} (A_{\mathcal{I}}^T A_{\mathcal{I}})^{-1}\succ 0.
% \end{align}

\emph{Stable Condition \eqref{stable_con_1}}: Note that $s A^T_j A_{\mathcal{I}}$ is a rank-1 matrix. From Lemma \ref{nonsymmetric_positve_definite}, we have
\begin{align}
    & A_{\mathcal{I}}^T A_{\mathcal{I}} - \gamma \text{sign}(-A_j^T(A_{\mathcal{I}} x_{\mathcal{I}}^* -y)) s A^T_j A_{\mathcal{I}} \succ 0  \nonumber \\ 
    \implies & 1 - \gamma \text{sign}(-A_j^T(A_{\mathcal{I}} x_{\mathcal{I}}^* -y)) A^T_j A_{\mathcal{I}} (A_{\mathcal{I}}^T A_{\mathcal{I}})^{-1}s > 0. \label{stable_con_3}
\end{align}
% Note that $s A^T_j A_{\mathcal{I}} (A_{\mathcal{I}}^T A_{\mathcal{I}})^{-1}$ is a rank-1 matrix, and the only eigenvalue is $A^T_j A_{\mathcal{I}} (A_{\mathcal{I}}^T A_{\mathcal{I}})^{-1}s $. 
Recall \eqref{median} \eqref{variable_a} defined in the previous subsection, we require
\begin{align*}
    & 1 - \gamma \text{sign}(-A_j^T(A_{\mathcal{I}} x_{\mathcal{I}}^* -y)) A^T_j A_{\mathcal{I}} (A_{\mathcal{I}}^T A_{\mathcal{I}})^{-1}s > 0 \\
    \implies & 1 - \gamma \text{sign}(a \lambda + b) a > 0 \\
    \implies & 1 - \frac{\lambda}{|a \lambda + b|} \text{sign}(a \lambda + b) a > 0 \\
    \implies & \frac{b}{a \lambda + b} > 0. 
\end{align*}
Under this condition ($b(a\lambda + b) > 0$), the function $\gamma(\lambda)$ is always increasing because $\gamma'(\lambda) > 0$, as discussed in Proposition \ref{multiple_fixed_points}.

\emph{Stable Condition \eqref{stable_con_2}}: This condition is equivalent to
\begin{align*}
     & 2I \succ \mu A_{\mathcal{I}}^T A_{\mathcal{I}} - \gamma \mu \text{sign}(-A_j^T(A_{\mathcal{I}} x_{\mathcal{I}}^* -y)) s A^T_j A_{\mathcal{I}} \\
     \implies & 2 (A_{\mathcal{I}}^T A_{\mathcal{I}} \! -\! \gamma \text{sign}(-A_j^T(A_{\mathcal{I}} x_{\mathcal{I}}^* -y)) s A^T_j A_{\mathcal{I}})^{-1} \succ \mu I
\end{align*}
When Stable Condition \eqref{stable_con_1} is satisfied, we choose $\mu$ to be the reciprocal of its largest eigenvalue. Because the rank-1 update matrix is nonsymmetric, it is not easy to predict the behavior of the eigenvalues. Consider the following condition 
\begin{align*}
     &2 \left(I - \gamma \text{sign}(-A_j^T(A_{\mathcal{I}} x_{\mathcal{I}}^* -y)) s A^T_j A_{\mathcal{I}} (A_{\mathcal{I}}^T A_{\mathcal{I}})^{-1} \right)^{-1} \\ 
     & \quad \quad \quad \quad \quad \quad \quad \quad \quad \quad \quad \quad \quad \quad \quad (A_{\mathcal{I}}^T A_{\mathcal{I}})^{-1} \succ \mu I,
\end{align*}
If the matrix in the inverse bracket only has two eigenvalues (not necessarily, but most likely), that is $1$ and $\frac{b}{a \lambda + b}$. To make the above condition hold, we need to set 
\begin{align} \label{mu_selection}
    \mu < \frac{2 \min(1, 1 + a\lambda/b)}{\|A_{\mathcal{I}}\|_2^2}.
\end{align}
\end{IEEEproof}

\begin{figure}[t!]
    \centering
    \includegraphics[width=0.5\textwidth]{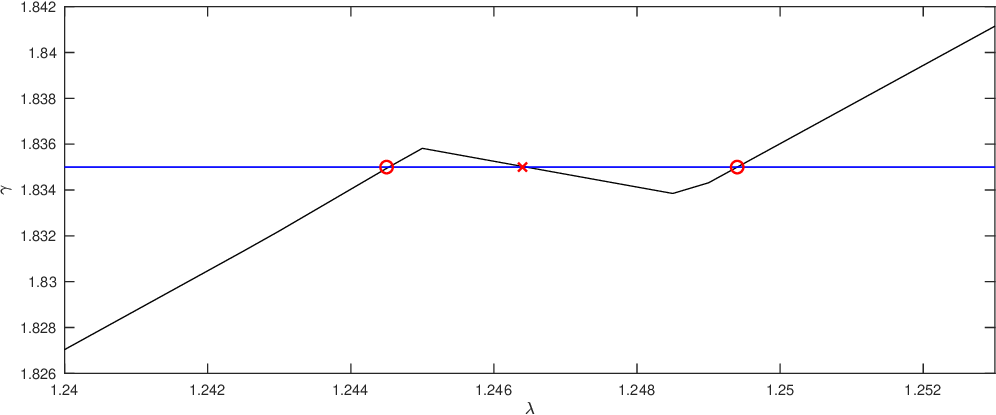}
    \caption{The stable points (marked in $\circ$) and unstable points (marked in $\times$) that satisfy the fixed point condition. \label{stable_unstable}}
\end{figure}

\begin{remark}
The upper bound of the step size $\mu$ in \eqref{mu_selection} is tight. However, it depends on the fixed point, because it involves with $a\lambda/b$ and $A_{\mathcal{I}}$. It is difficult to have a precise lower bound of $1 + a\lambda/b$ due to the same reason. Intuitively, $1 + a\lambda/b$ is bounded aways from 0, because $a\lambda + b = -A_j^T(A_{\mathcal{I}} x_{\mathcal{I}}^* -y)$ is the signed $\text{median}(|- A^T(A \bar{x}(\lambda) - y)|)$, and thus is never 0; and by definition of $b$ in \eqref{variable_b}, the projection of $y$ onto the range of $A_{\mathcal{I}}$, is also bounded away from infinity. Hence, the upper bound is always a positive number. From our experience, it is safe to set $\mu < 2/\|A\|_2^2$.
\end{remark}

\begin{figure}[t!]
    \centering
    \includegraphics[width=0.45\textwidth]{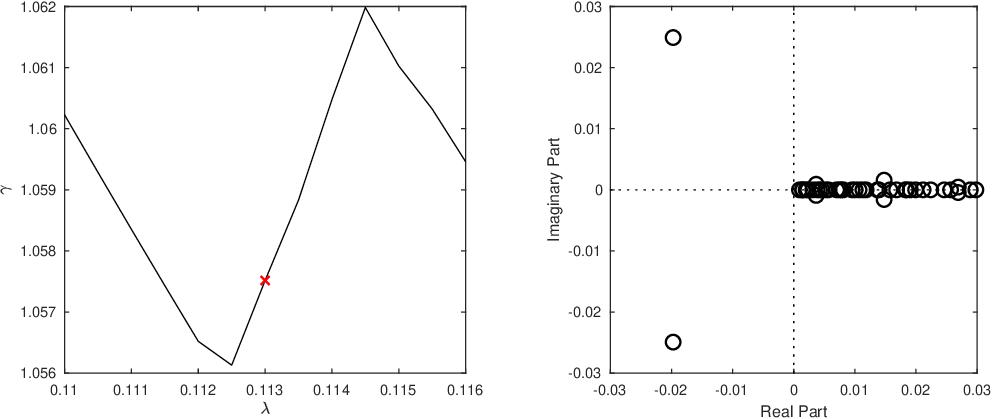}
    \caption{A LASSO solution as an unstable points (marked in $\times$), while $\gamma(\lambda)$ is increasing. $\frac{b}{a \lambda + b} > 0$ but $A_{\mathcal{I}}^T A_{\mathcal{I}} - \gamma \text{sign}(-A_j^T(A_{\mathcal{I}} x_{\mathcal{I}}^* -y)) s A^T_j A_{\mathcal{I}}$ has a pair of complex eigenvalues with negative real part. \label{unstable_point}}
\end{figure}
\begin{figure}[!]
    \centering
    \includegraphics[width=0.45\textwidth]{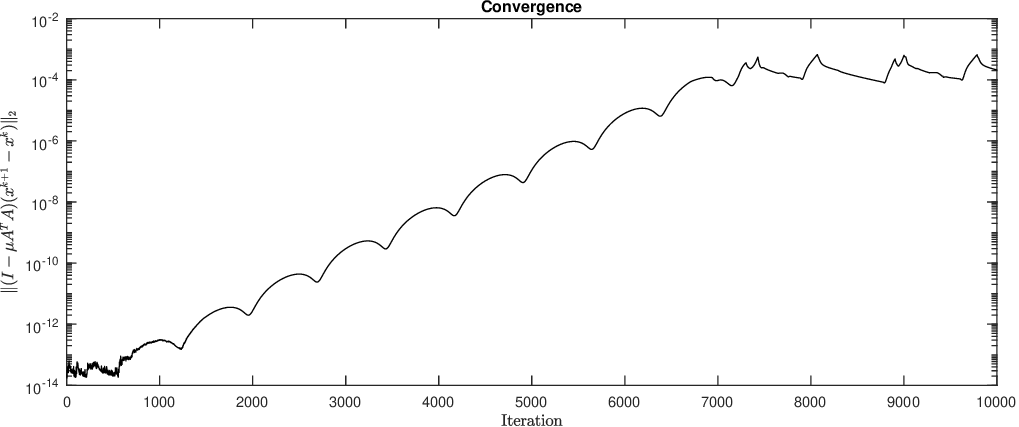}
    \caption{The divergence of the proposed algorithm when initialized on the unstable LASSO solution in Fig. \ref{unstable_point}. \label{unstable_converge}}
\end{figure}

An illustration of the stable and unstable candidate fixed points from Proposition \ref{prop_stability} is shown in Fig. \ref{stable_unstable}. However, Proposition \ref{prop_stability} is only a necessary condition, because Lemma \ref{nonsymmetric_positve_definite} is a necessary condition. That is, if a LASSO solution of $\lambda$ is a stable fixed point, then $\gamma(\lambda)$ is increasing. We can find scenarios where $\gamma(\lambda)$ is increasing, and the corresponding LASSO solution is unstable. Fig. \ref{unstable_point} \ref{unstable_converge} shows such a case.

\section{Convergence Analysis of Adaptive ISTA with the MAD}

As we established the properties of the fixed points of \eqref{AISTA} in the last section, the natural next step is to investigate its convergence behavior. The convergence analysis of adaptive ISTA with any thresholding strategy, especially building global convergence guarantees, is difficult, due to the constantly changing thresholds from iterations as the algorithm progresses. The adaptive ISTA often fails to converge when the thresholds diverge. Nevertheless, the adaptive ISTA with the MAD thresholding strategy empirically appears to have a pretty stable convergence behavior, regardless of the initialization and the choice of step size. Intuitively speaking, it is because the MAD is a robust measure of the variability of a bunch of samples, and does not change too much from perturbations, which helps the convergence. 

In this section, we give theoretical analyses of the convergence behavior of adaptive ISTA with the MAD. The center of the analysis is on a matrix recurrence inspired by \cite{tao2016local}. The matrix recurrence largely simplifies the analysis, and convergence or divergence of iterations can be determined by spectral analysis of the matrix recurrence.  

\begin{definition}
Assume the median of the positive vector $|z|$ occurs at index $j$, the adaptive thresholding operator $T:\mathbb{R^N} \to \mathbb{R}^N$ in \eqref{AISTA} with the MAD has the matrix multiplication formula
\begin{align}
    T(z) = (D^2 - \gamma \, \text{sign}(z_j) \, d \, e_j^T ) \, z,
\end{align}
where $d \in \mathbb{R}^N$ is defined element-wise as
\begin{align}
    d_i = \text{sign}(T(z_i)) = \begin{cases}
    +1 & z_i > \gamma |z_j|,\\
    0 & |z_i| < \gamma |z_j|, \\
    -1 & z_i < - \gamma |z_j|,
    \end{cases}
\end{align}
$D = \text{diag}(d)$, and thus $D^2$ is a diagonal matrix for support selection, with $1$ for $T(z_i) \not = 0$, and $0$ for $T(z_i) = 0$. And $e_j$ is the canonical basis for median index selection where the element at index $j$ is 1 and otherwise 0. 
\end{definition}

\begin{remark}
    The matrix $D^2 - \gamma \, \text{sign}(z_j) \, d \, e_j^T$ has explicit formula 
    \begin{align}
    &D^2 - \gamma \, \text{sign}(z_j) \, d \, e_j^T \\ \nonumber
    & = \begin{bmatrix}
        1 & & \cdots & -\gamma \text{sign}(z_1)\text{sign}(z_j) &  & \\
          &0& \cdots & 0 &  & \\
          & & \ddots & \vdots &  & \\ 
      & &  & \vdots & \ddots &  \\
         & &  & -\gamma \text{sign}(z_N)\text{sign}(z_j) & \cdots & 1 \\    
    \end{bmatrix}.
\end{align}
\end{remark}

\begin{proposition}
The adaptive ISTA with the MAD thresholding strategy \eqref{AISTA} is equivalent to the following matrix recurrence
\begin{align} \label{matrix_recurrence}
    \begin{bmatrix} z^{k+1} \\ 1 \end{bmatrix} = 
     \begin{bmatrix}
        B^k & \mu A^T y \\ \underline{0}^T & 1
    \end{bmatrix} \begin{bmatrix} z^k \\ 1 \end{bmatrix},
\end{align}
where 
\begin{align} \label{flag_matrix}
    B^k = (I - \mu A^T A)\left( \left(D^k\right)^2 - \gamma \, \text{sign}(z^k_j) \, d^k \, \left(e^k_j\right)^T \right),
\end{align}
and $\underline{0}$ is a zero column vector.
\end{proposition}

\begin{IEEEproof}

Let $z^k = x^k - \mu A^T(Ax^k - y)$, then 
\begin{align*}
 & x^{k+1} = T(x^k - \mu A^T(Ax^k - y )) \\
 \implies & x^{k+1} - \mu A^T(Ax^{k+1} - y ) \\
 & \quad \quad  = T(x^k - \mu A^T(Ax^k - y )) - \mu A^T(Ax^{k+1} - y ) \\ 
 \implies & x^{k+1} - \mu A^T(Ax^{k+1} - y ) \\
 & \quad \quad= (I - \mu A^T A)\,T(x^k - \mu A^T(Ax^k - y )) + \mu A^T y  \\ 
 \implies & z^{k+1} = (I - \mu A^T A)\,T(z^k) + \mu A^T y. 
\end{align*}
By the matrix multiplication formula of $T$, we have
\begin{align*}
    z^{k+1} & = (I - \mu A^T A)\left( \left(D^k\right)^2 - \gamma \, \text{sign}(z^k_j) \, d^k \, \left(e^k_j\right)^T \right) \, z^k \\
    & \quad \quad \quad \quad \quad \quad \quad \quad \quad \quad \quad \quad \quad \quad \quad \quad  + \mu A^T y,
\end{align*}
which can be written as the matrix recurrence in \eqref{matrix_recurrence}. 
\end{IEEEproof}

\begin{remark}
The matrix recurrence in \eqref{matrix_recurrence} is not a practical algorithm, and is for analysis only. We need the actual algorithm \eqref{AISTA} to decide the median index $j$, the sign of $T(z)$ in $d^k$ and the support $(D^k)^2$.
\end{remark}

It is clear that the block matrix in \eqref{matrix_recurrence} determines whether the update at iteration $k$ forms a contraction and progresses to convergence. Since the block matrix automatically has one eigenvalue equal to 1, the eigenvalues of $B^k$ have to be strictly within the unit circle for the matrix recurrence to be contractive. 

\subsection{Local Linear Convergence}

In this subsection, we perform spectral analysis on $B^k$ and show that the algorithm \eqref{AISTA} converges linearly around a neighborhood of a stable fixed point, the same behavior as the regular ISTA \eqref{ISTA} as proven in \cite{tao2016local}.

\begin{proposition}
Around the neighborhood of a stable fixed point $x^*$ of the algorithm \eqref{AISTA}, at iteration $k$ where $T(z^k)$ has the same median index $j$, support and sign $d_{\mathcal{I}}^k = s$ as that of the fixed point, the algorithm converges linearly to $x^*$.   
\end{proposition}

\begin{IEEEproof}
First, note that the columns of $B^k$ are non-zeros only on the support $\mathcal{I}$ and index $j$. Since rearranging rows and columns in the same order does not change the eigenvalue, we investigate the following block matrix $\bar{B}^k$,
    \begin{align*}
        \bar{B}^k &\! =\! \begin{bmatrix} I - \mu A_{\mathcal{I}}^T A_{\mathcal{I}} & 0 \\ - \mu A_{-\mathcal{I}}^T A_{\mathcal{I}} & 0 \end{bmatrix} \left( I  - \gamma \, \text{sign}(z^k_j) \, \begin{bmatrix} s \\ \underline{0} \end{bmatrix} \, \left(e^k_{\bar{j}}\right)^T \right) \\
        & \!=\! \begin{bmatrix} I \!-\! \mu A_{\mathcal{I}}^T A_{\mathcal{I}} & 0 \\ - \mu A_{-\mathcal{I}}^T A_{\mathcal{I}} & 0 \end{bmatrix} \! -\! \gamma \, \text{sign}(z^k_j)\! \begin{bmatrix} (I \!-\! \mu A_{\mathcal{I}}^T A_{\mathcal{I}})s \\ - \mu A_{-\mathcal{I}}^T A_{\mathcal{I}} s \end{bmatrix} \! \left(e^k_{\bar{j}}\right)^T 
    \end{align*}
where we switched all the support $\mathcal{I}$ to the first $|\mathcal{I}|$ rows and columns, and the median index $\bar{j}$ is among the off-support indices. To form a contraction, we require 
\begin{align*}
    I \succ \bar{B}^k \succ -I, 
\end{align*}
that is
\begin{align} 
 & \begin{bmatrix}  \mu A_{\mathcal{I}}^T A_{\mathcal{I}} & 0 \\ \mu A_{-\mathcal{I}}^T A_{\mathcal{I}} & I \end{bmatrix} \! +\! \gamma \, \text{sign}(z^k_j) \! \begin{bmatrix} (I \!-\! \mu A_{\mathcal{I}}^T A_{\mathcal{I}})s \\ - \mu A_{-\mathcal{I}}^T A_{\mathcal{I}} s \end{bmatrix} \! \left(e^k_{\bar{j}}\right)^{\!T\!} \!\succ \! 0, \label{local_converge_1} \\    
 & \begin{bmatrix} 2I \! -\! \mu A_{\mathcal{I}}^T A_{\mathcal{I}} & 0 \\ - \mu A_{-\mathcal{I}}^T A_{\mathcal{I}} & I \end{bmatrix} \! - \! \gamma \, \text{sign}(z^k_j) \! \begin{bmatrix} (I \!-\! \mu A_{\mathcal{I}}^T A_{\mathcal{I}})s \\ - \mu A_{-\mathcal{I}}^T A_{\mathcal{I}} s \end{bmatrix} \! \left(e^k_{\bar{j}}\right)^{\!T\!} \! \succ \! 0. \label{local_converge_2}
\end{align}
The two conditions are similar to \eqref{stable_con_1}\eqref{stable_con_2}, therefore we can use the same technique for analysis. By Lemma \ref{nonsymmetric_positve_definite},
\begin{align*}
    &\text{det}\left(\begin{bmatrix}  \mu A_{\mathcal{I}}^T A_{\mathcal{I}} & 0 \\ \mu A_{-\mathcal{I}}^T A_{\mathcal{I}} & I \end{bmatrix} \! +\! \gamma \, \text{sign}(z^k_j) \! \begin{bmatrix} (I \!-\! \mu A_{\mathcal{I}}^T A_{\mathcal{I}})s \\ - \mu A_{-\mathcal{I}}^T A_{\mathcal{I}} s \end{bmatrix} \! \left(e^k_{\bar{j}}\right)^{\!T\!} \right)  \\
  \!=\! & \text{det}\left(\begin{bmatrix}  \mu A_{\mathcal{I}}^T A_{\mathcal{I}} & 0 \\ \mu A_{-\mathcal{I}}^T A_{\mathcal{I}} & I \end{bmatrix} \right) \\
  &\left(\!1 \! +\! \gamma \, \text{sign}(z^k_j) \!\left(e^k_{\bar{j}}\right)^{\!T\!} \! {\begin{bmatrix}  \mu A_{\mathcal{I}}^T A_{\mathcal{I}} & 0 \\ \mu A_{-\mathcal{I}}^T A_{\mathcal{I}} & I \end{bmatrix}}^{-1} \!\begin{bmatrix} (I\! - \!\mu A_{\mathcal{I}}^T A_{\mathcal{I}})s \\ - \mu A_{-\mathcal{I}}^T A_{\mathcal{I}} s \end{bmatrix} \!\right)\!,
\end{align*}
similarly, we require the term in the bracket to be positive, 
\begin{align*}
    1 \! +\! \gamma \, \text{sign}(z^k_j) \!\left(e^k_{\bar{j}}\right)^{\!T\!} \! {\begin{bmatrix}  \mu A_{\mathcal{I}}^T A_{\mathcal{I}} & 0 \\ \mu A_{-\mathcal{I}}^T A_{\mathcal{I}} & I \end{bmatrix}}^{-1} \!\begin{bmatrix} (I\! - \!\mu A_{\mathcal{I}}^T A_{\mathcal{I}})s \\ - \mu A_{-\mathcal{I}}^T A_{\mathcal{I}} s \end{bmatrix}  > 0.
\end{align*}
By Schur complement, we have a closed formula for the inverse of a block matrix, 
\begin{align*}
     & \!1\! +\! \gamma \, \text{sign}(z^k_j)\! \left(e^k_{\bar{j}}\right)^{\!T\!} \\
     & \quad \quad \begin{bmatrix}  (\mu A_{\mathcal{I}}^T A_{\mathcal{I}})^{-1} & 0 \\ -A_{-\mathcal{I}}^T A_{\mathcal{I}} (A_{\mathcal{I}}^T A_{\mathcal{I}})^{-1} & I \end{bmatrix} \begin{bmatrix} (I - \mu A_{\mathcal{I}}^T A_{\mathcal{I}})s \\ - \mu A_{-\mathcal{I}}^T A_{\mathcal{I}} s \end{bmatrix} > 0 \\
    \implies &\! 1\! +\! \gamma \, \text{sign}(z^k_j) \!\left(e^k_{\bar{j}}\right)^T \! \begin{bmatrix} (\mu A_{\mathcal{I}}^T A_{\mathcal{I}})^{-1}(I - \mu A_{\mathcal{I}}^T A_{\mathcal{I}})s \\ -A_{-\mathcal{I}}^T A_{\mathcal{I}} (A_{\mathcal{I}}^T A_{\mathcal{I}})^{-1} s \end{bmatrix} > 0\\
    \implies & 1 - \gamma \, \text{sign}(z^k_j) \, A_j^T A_{\mathcal{I}} (A_{\mathcal{I}}^T A_{\mathcal{I}})^{-1} s > 0. 
\end{align*}
By definition, $z_j^k = x_j^k - \mu [A^T(Ax^k - y)]_j$. And $x_j^k = 0$ since it is not in the support $\mathcal{I}$, $z_j^k = - \mu [A^T(Ax^k - y)]_j = - \mu A_j^T(Ax_{\mathcal{I}}^k - y)$, we require
\begin{align*}
    1 - \gamma \, \text{sign}(- \mu A_j^T(Ax_{\mathcal{I}}^k - y)) \, A_j^T A_{\mathcal{I}} (A_{\mathcal{I}}^T A_{\mathcal{I}})^{-1} s > 0,
\end{align*}
which is exactly the same as \eqref{stable_con_3}. The analysis on \eqref{local_converge_2} can be done in the same way. 

Thus when \eqref{local_converge_1}\eqref{local_converge_2} holds, $B^k$ has all eigenvalues strictly with the unit circle, and the matrix recurrence \eqref{matrix_recurrence} reduces to the power method on a constant matrix, which converges at a linear rate \cite{tao2016local}. 
\end{IEEEproof}

\subsection{Global Convergence Behavior}

In the last subsection, we prove that the proposed algorithm converges linearly around a stable fixed point, in the last stages of the proposed algorithm where the matrix $B^k$ in \eqref{matrix_recurrence} does not change. However, there is no guarantee that $I \succ B^k \succ -I$ throughout the progress of the algorithm, and it is hard to do analyses on how $B^k$ could change between iterations because the algorithm (more specifically, the soft-thresholding operator with the MAD) decides the support and median index itself. To obtain a convergence guarantee beyond the one we presented, we need to make further assumptions on the problem setups, which is beyond the scope of this paper. 

\begin{figure}[t]
    \centering
    \includegraphics[width=0.45\textwidth]{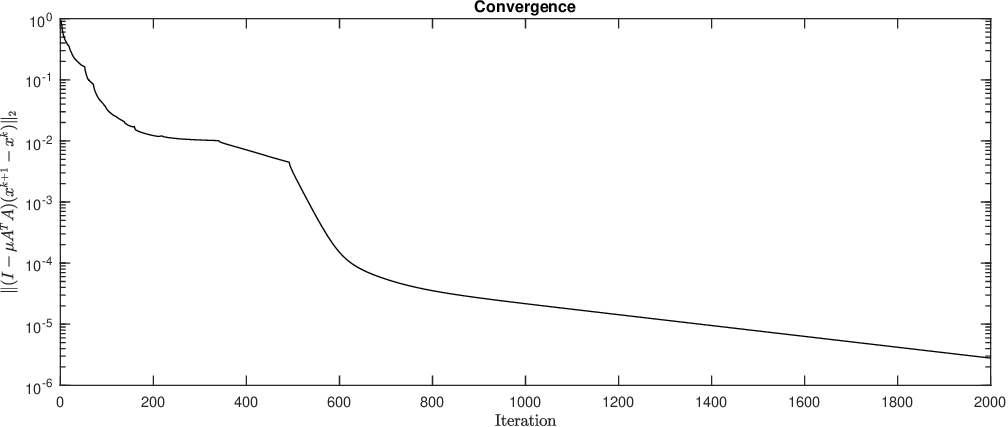}
    \caption{The piecewise convergence behavior of the proposed algorithm. \label{converge}}
\end{figure}

Nevertheless, we observe that the convergence behaves in a piecewise way, because the support, median index and $B^k$ do not change frequently (especially when the iteration gets closer to a fixed point). That is, it either converges at a fixed linear rate (because $B^k$ stays the same), or diverges at a fixed linear rate. Fig. \ref{converge} shows the piecewise behavior, which is prominent at iterations 200-350, 350-500, and 500-2000.  

\section{Numerical Experiments}

\begin{figure}[t!]
    \centering
    \includegraphics[width=0.5\textwidth]{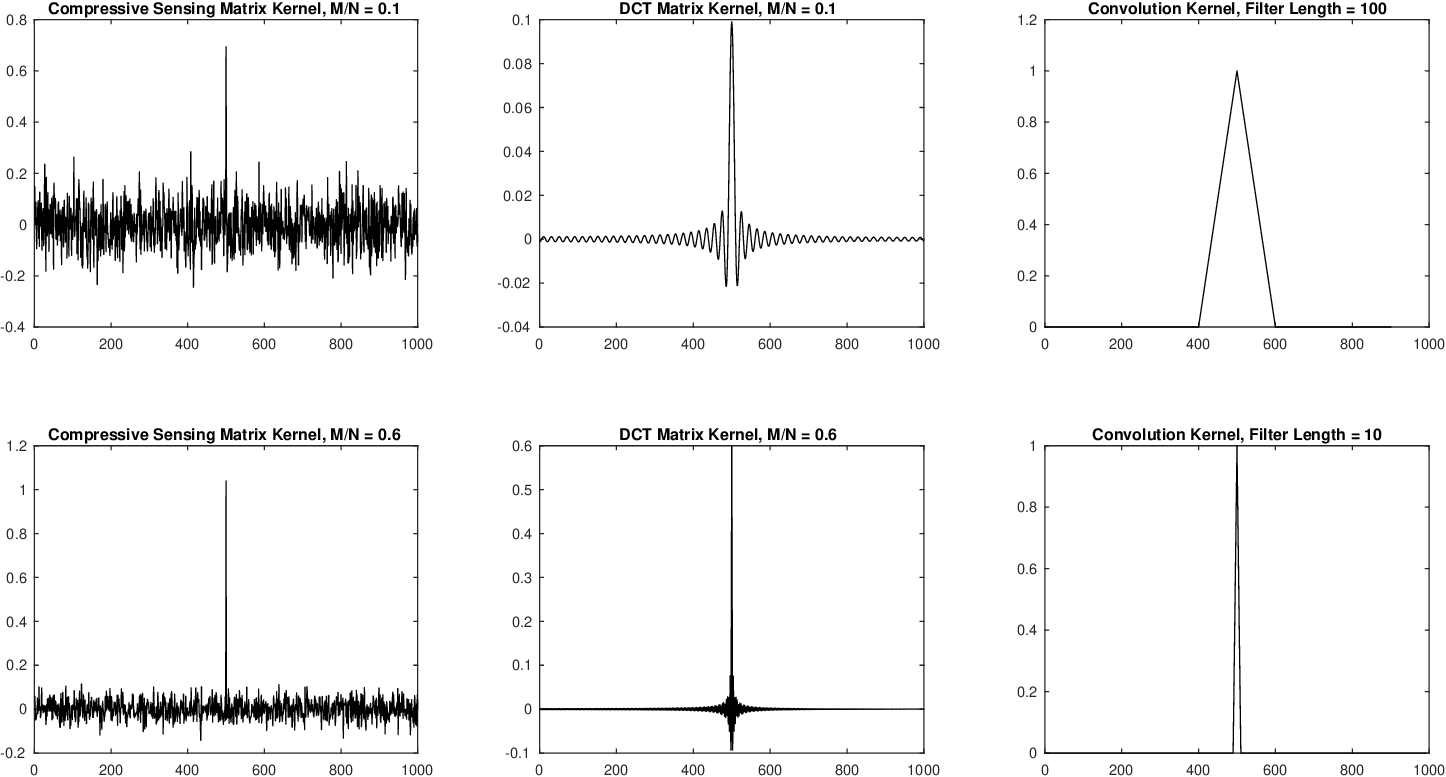}
    \caption{Matrix kernel of $A^T A$ in different settings. \label{matrix_kernel}}
\end{figure}

We show the performance of three different types of inverse problems in this section. The problem setups and their relevant parameters are
\begin{enumerate}
\item Compressive Sensing: $A \in \mathbb{R}^{M \times N}, A_{i,j} \sim \mathcal{N}(0, 1/M)$   
\begin{itemize}
\item Problem Size $N$, Undersampling Ratio $M/N$,  
\end{itemize}

\item DCT: $A \in \mathbb{R}^{M \times N}, A$ is the truncated DCT matrix (first $M$ rows)   
\begin{itemize}
\item Problem Size $N$, Undersampling Ratio $M/N$,  
\end{itemize}

\item Deconvolution: $A \in \mathbb{R}^{M \times N}, A$ is the convolutional matrix with moving average filter  
\begin{itemize}
\item Problem Size $N$, Filter/Kernel Length $L$.  
\end{itemize}
\end{enumerate}
The ground truth sparse signal $x_0$ is drawn i.i.d. from the Bernoulli-Gaussian distribution with sparsity level $\rho$ (i.e. $p_X(x_j) = (1 - \rho)\delta(x_j) + \rho \mathcal{N}(x_j; 0, 1/\rho) \forall j$, where $\delta(\cdot)$ denote the Dirac delta distribution).  
The three problem setups are not fundamentally different, because the kernels $A^T A$ induced from the measurement matrix behave similarly, illustrated in Fig. \ref{matrix_kernel}. That is
\begin{enumerate}
\item Compressive Sensing: $\mathbb{E} [A^T A] = I$. As $M/N: 0 \to 1$, $A^T A$ becomes closer to identity. That is the reason why the compressive sensing problem behaves like a sparse denoising problem \cite{donoho2009message}.     

\item DCT: The kernel $A^T A$ is a sinc function. As $M/N: 0 \to 1$, the function decays faster to 0, and $A^T A$ becomes closer to identity.   

\item Deconvolution: The kernel $A^T A$ is a hat function. As the filter length becomes smaller, the function decays faster to 0, and $A^T A$ becomes closer to identity. 
\end{enumerate}
Intuitively speaking, as the kernels become more localized and closer to identity, the corresponding sparse linear inverse problems become closer to the sparse denoising problem, and thus easier. Nevertheless, it is way more challenging to build theoretical claims upon such intuition for the DCT and deconvolution problems with deterministic kernels.   

\begin{figure}[!]
    \centering
    \includegraphics[width=0.45\textwidth]{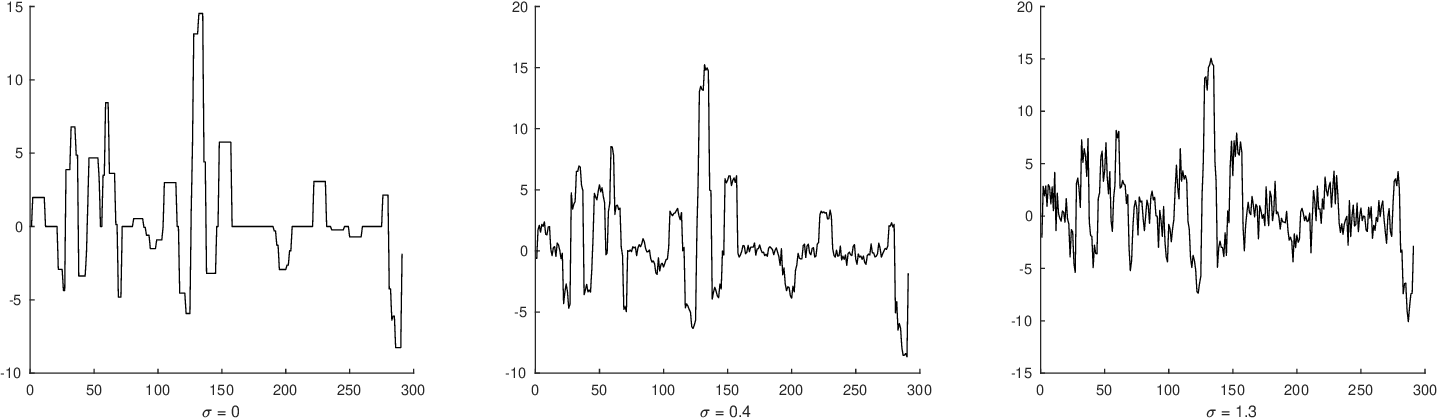}
    \includegraphics[width=0.45\textwidth]{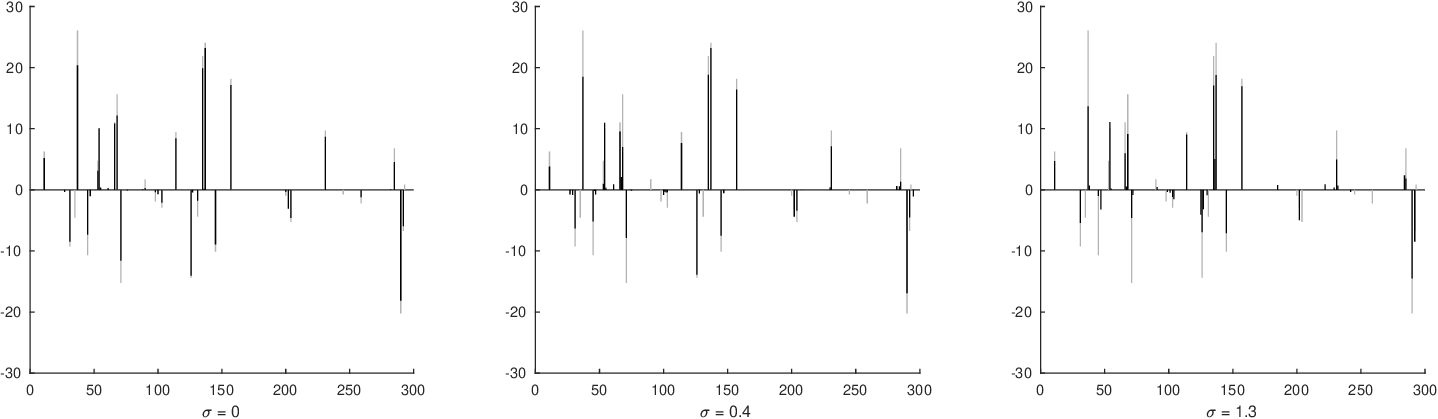}
    \caption{Example of sparse deconvolution problem, noisy observations corrupted by noise with different noise levels, and the corresponding estimation with ISTA with the MAD. \label{deconv_example}}
\end{figure}

\begin{figure*}[t!]
    \centering
    \includegraphics[angle=0,width=0.3\textwidth]{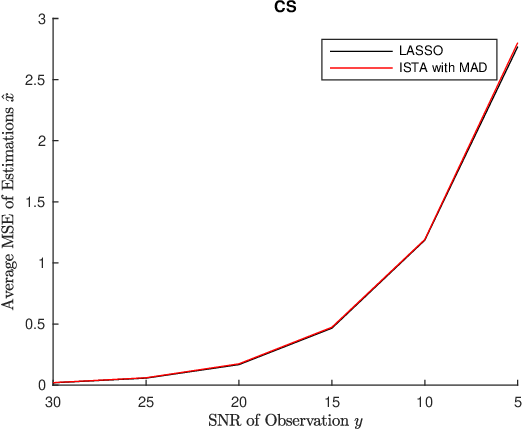}
    \includegraphics[angle=0,width=0.3\textwidth]{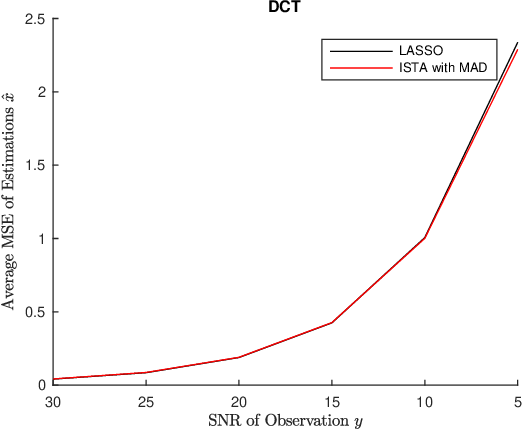}
    \includegraphics[angle=0,width=0.3\textwidth]{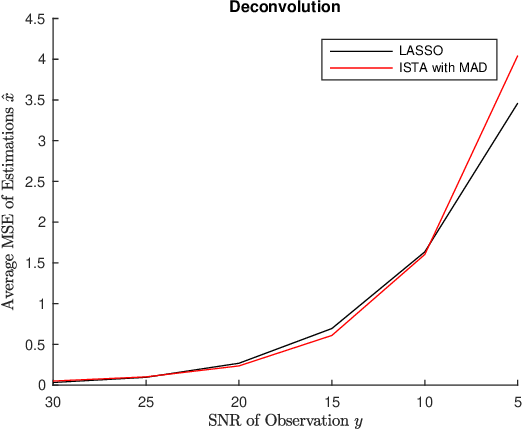}\\
    \vspace{0.3cm}
\includegraphics[angle=0,width=0.3\textwidth]{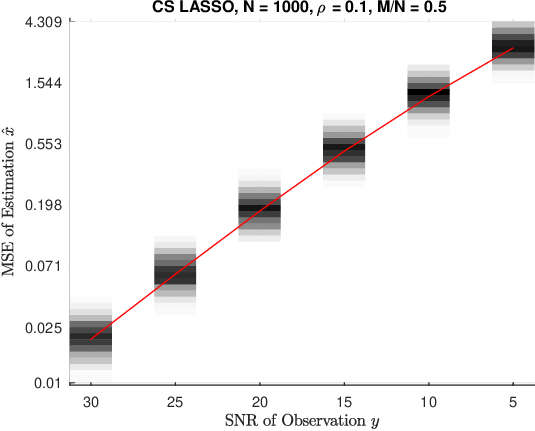} 
\includegraphics[angle=0,width=0.3\textwidth]{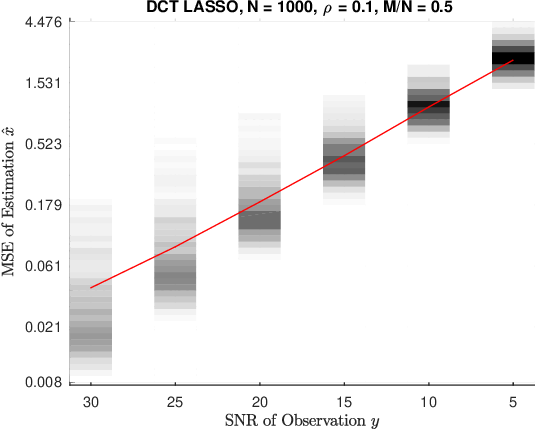}
\includegraphics[angle=0,width=0.3\textwidth]{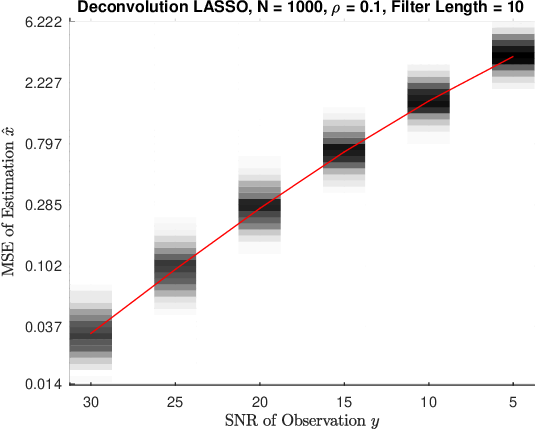}\\
\vspace{0.3cm}
\includegraphics[angle=0,width=0.3\textwidth]{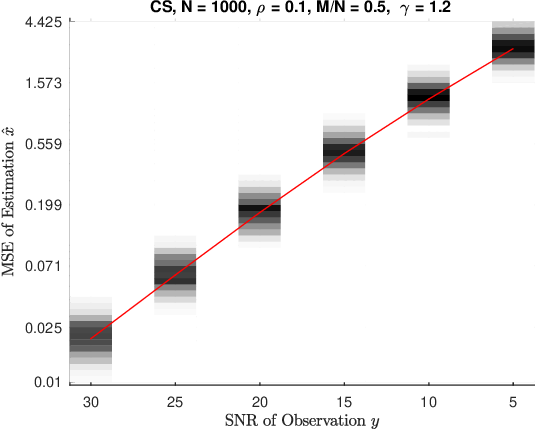} 
\includegraphics[angle=0,width=0.3\textwidth]{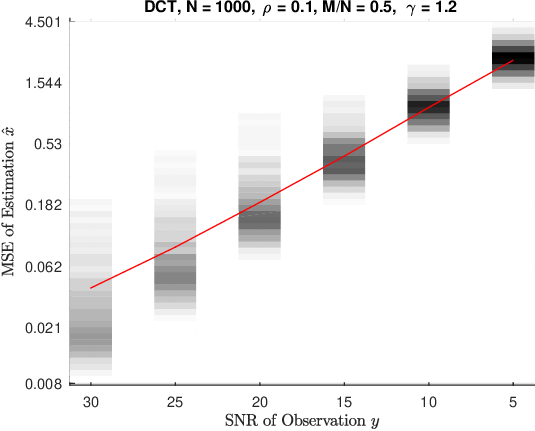} 
\includegraphics[angle=0,width=0.3\textwidth]{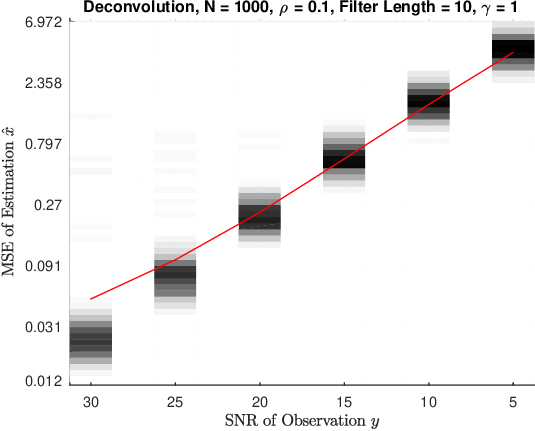}
\caption{Performance comparison between ISTA with MAD and LASSO solution with known noise level. \label{fig_MAD_LAsso}}
\end{figure*}

Despite the theoretical difficulty, we can verify some good properties with the numerical experiments.  The goal of ISTA with the MAD is to "automatically" tune $\lambda$ for the LASSO problem in various settings of noise levels.
Fig. \ref{deconv_example} shows an example of the deconvolution problem, with observation corrupted by noise with different noise levels. The estimations from ISTA with the MAD (with $\gamma = 1.2 \times 1.4826$ fixed for all noise levels) are consistently good among different noise levels. 
This indicates that ISTA with the MAD can implicitly find an appropriate $\lambda$ for the LASSO problem. 
The corresponding $\lambda$s for the three estimations are $1.045, 2.073, 2.554$. Note that we have to try out significantly many $\lambda$s for each noise level if we have to tune $\lambda$ of the LASSO problem, which is computationally expensive.

We will show and compare the performance of ISTA with the MAD against other methods in the following subsections. Specifically, we obtain different estimations from 1000 realizations\footnote{The $x_0$ and $w$ are generated according to their distribution for each realization. For the compressive sensing problem, the measurement matrix $A$ is randomly generated for each realization.} under various observation SNR ($30 \to 5$, corresponding to noise level from low to high), compute the MSE between the estimations and the ground truth, and compare the average MSE and the distribution of the MSEs by the histogram.

\subsection{ISTA with the MAD \& LASSO Solutions with Known Noise Level}

The ISTA with the MAD is considered a "tuning-free" method for selecting appropriate $\lambda$ for the LASSO problem in various noise levels. To verify that ISTA with the MAD can implicitly find a $\lambda$ related to noise level, we compare its performance against LASSO solutions, where the $\lambda$ is set as scaled noise level $\sigma$ ($1.2 \sigma$ for compressive sensing and DCT problems, $\sigma$ for deconvolution problem), with $\sigma$ known as prior knowledge. For ISTA with the MAD, we set $\gamma = 1.2 \times 1.4826$ for compressive sensing and DCT problems, and $\gamma = 1.4826$ for deconvolution problem. 

Fig. \ref{fig_MAD_LAsso} shows the comparison. The first row compares the average of MSEs of the two methods from 1000 realizations on the three problems, with the average of MSE ($y$-axis) plotted on a linear scale. For the compressive sensing and the DCT problem, the performances of the two methods are identical in terms of average MSE. For the deconvolution problem, the two methods are almost identical, except in the really high noise level regime (observation SNR is 5). 

\begin{figure*}[t!]
    \centering
    \includegraphics[angle=0,width=0.3\textwidth]{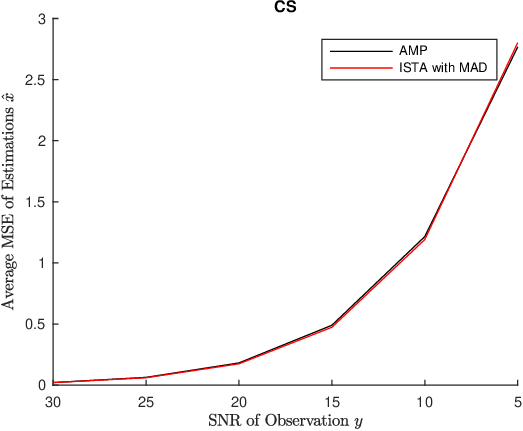}
\includegraphics[angle=0,width=0.3\textwidth]{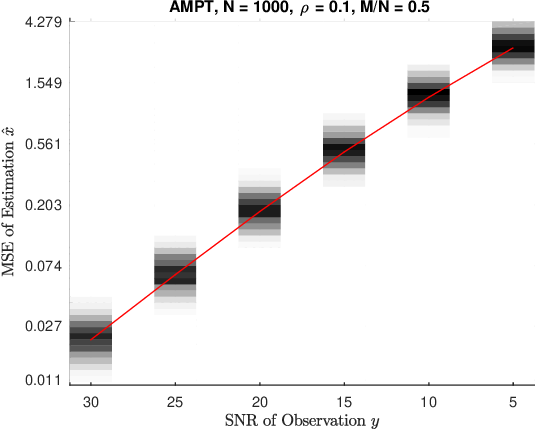} 
\includegraphics[angle=0,width=0.3\textwidth]{images/adaptive_randn_1000_01_05_1dot2.eps} 
\caption{Performance comparison between ISTA with MAD and AMP. \label{fig_MAD_AMP}}
\end{figure*}

The second and third row shows the histogram of the MSEs of LASSO solutions and ISTA with the MAD, respectively, with MSE ($y$-axis) plotted on a log scale\footnote{The observation SNR ($x$-axis) changes in a log scale. The more noise-corrupted the observations are, the more variability the estimations have, so do the MSEs. To clearly illustrate the distribution, especially in the small noise regime, a log scale is adopted.}. The histograms are represented as blocks on a gray color scale: the darker the block is, the more MSEs out of the 1000 realizations fall into the bin, and vice versa. For the compressive sensing problem, the MSEs distribute according to a normal distribution around the average, indicating the two methods behave identically well. For the DCT problem, the MSEs have a dispersed and irregular distribution, especially in the low noise regime; but the two methods have identical behavior, indicating the DCT problem is fundamentally challenging for both. For the deconvolution problem, the MSEs distribute according to a normal distribution, except for ISTA with the MAD in the low noise regime, where there are a few bad-behaving outliers.     

In short, the ISTA with the MAD can match the performance of the LASSO solutions with known noise levels, except in the challenging deconvolution problem in the really low and high noise regime. We can conclude that the ISTA with the MAD can implicitly detect the noise level and find a good estimations accordingly.

\subsection{ISTA with the MAD \& AMP}

AMP was introduced to solve the compressive sensing problem \cite{donoho2009message} and proven to be the optimal algorithm. The fixed points of the variant "Tunable AMP" correspond to LASSO solutions \cite{donoho2011noise} and adapt to various noise levels. However, AMP does not converge for the problems with deterministic measurement matrix $A$. 
Similar to the last subsection, we compare the ISTA with the MAD against the AMP for the compressive sensing problem. 

Fig. \ref{fig_MAD_AMP} shows the comparison. We can see that the two methods perform identically in terms of both average MSE and the distribution of MSEs.

\subsection{Asymptotic Consistency}

The asymptotic consistency is a desirable property of estimation methods in general. Specifically, to the sparse linear inverse problem, with all the parameters held fixed\footnote{Sparsity level $\rho$, observation SNR, undersampling ratio $M/N$ for compressive sensing and DCT problem, Filter Length for deconvolution problem.}, as the size of the problem $N$ increases indefinitely, the asymptotic consistent estimation should converge in probability. Intuitively speaking, as $N$ increases, the distributions of the estimations become more and more concentrated, and so do the distributions of MSEs. 

\begin{figure*}[t!]
    \centering
\includegraphics[angle=0,width=0.3\textwidth]{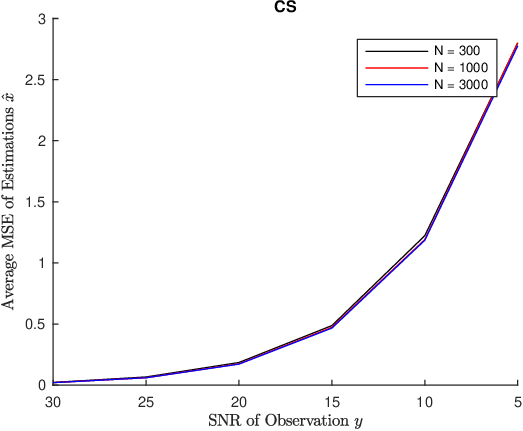}
\includegraphics[angle=0,width=0.3\textwidth]{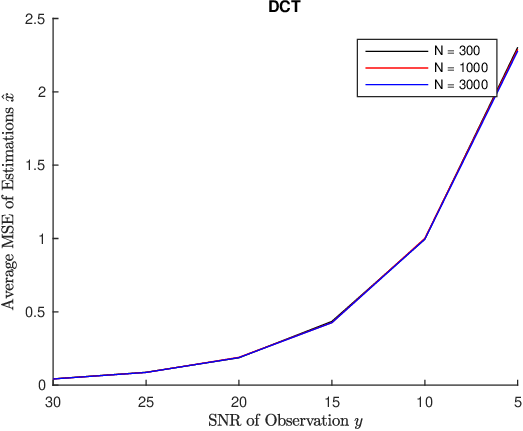}
\includegraphics[angle=0,width=0.3\textwidth]{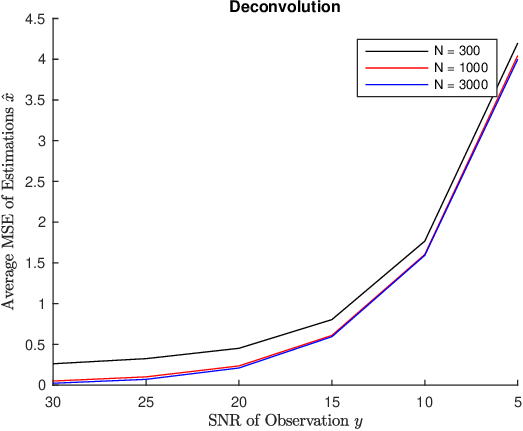} \\
    
\includegraphics[angle=0,width=0.3\textwidth]{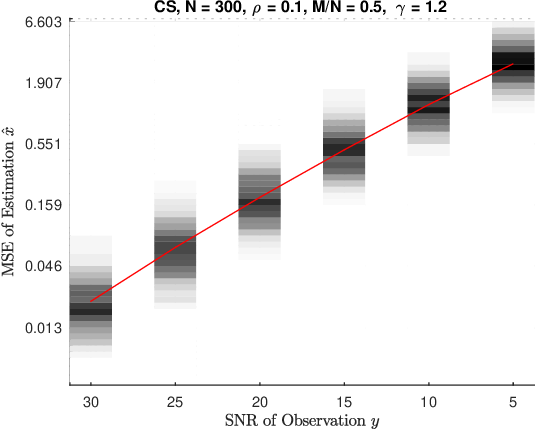}
\includegraphics[angle=0,width=0.3\textwidth]{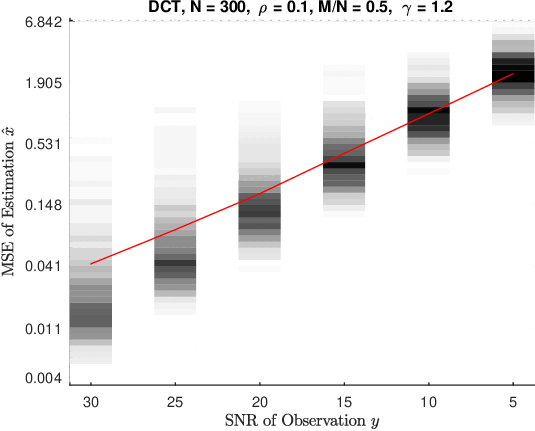}
\includegraphics[angle=0,width=0.3\textwidth]{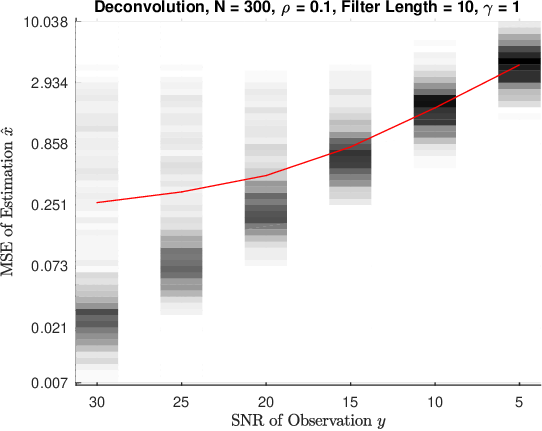} \\

\includegraphics[angle=0,width=0.3\textwidth]{images/adaptive_randn_1000_01_05_1dot2.eps}
\includegraphics[angle=0,width=0.3\textwidth]{images/adaptive_dct_1000_01_05_1dot2.eps}
\includegraphics[angle=0,width=0.3\textwidth]{images/adaptive_deconv_1000_01_05_1.eps} \\

\includegraphics[angle=0,width=0.3\textwidth]{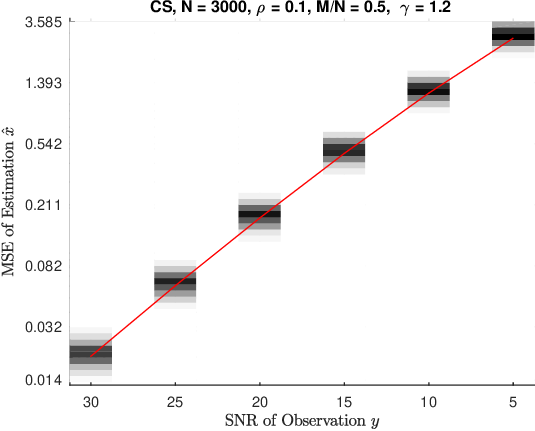}
\includegraphics[angle=0,width=0.3\textwidth]{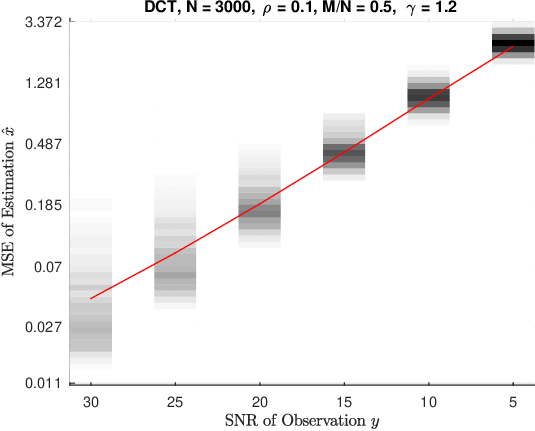} 
\includegraphics[angle=0,width=0.3\textwidth]{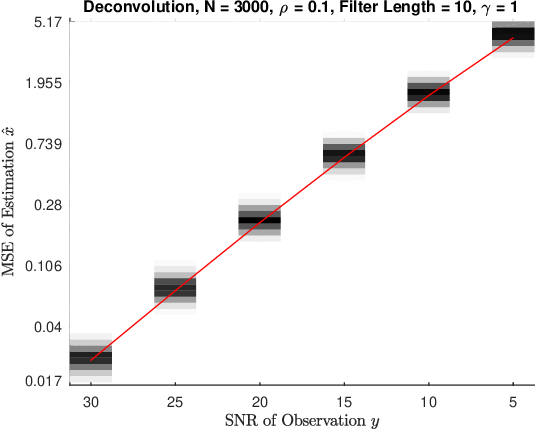} 

\caption{Asymptotic Behavior of ISTA with MAD. \label{fig_asymptotic}}
\end{figure*}

Fig. \ref{fig_asymptotic} provides numerical evidences that the estimations from ISTA with the MAD are asymptotic consistent. 
The first row illustrates the average of MSEs from 1000 realizations for the three problems at different problem sizes $N = 300, 1000, 3000$. For the compressive sensing and DCT problems, the average MSEs are identical for all $N$. For deconvolution problem, the average MSEs converge as $N$ grows (identical for $N = 1000, 3000$).  
The second to last rows show the distribution of the MSEs at problem sizes $N = 300, 1000, 3000$, respectively. For compressive sensing problem, the distributions of MSEs not only become more concentrated as $N$ increases, but they also follow a normal distribution (known as asymptotic normality). 
For DCT problem, the distributions of MSEs become more concentrated as $N$ increases, except in low noise regime (SNR = 30). However, the distributions are not as regular as those in the compressive sensing case. 
For the deconvolution problem, the distributions of MSEs become more concentrated as $N$ increases, yet the distributions for $N = 300$ have a completely different behavior than those for $N = 1000, 3000$.

% \subsection{Consistent Performance Among $\gamma$s}

\section{Conclusion \& Future Work}

In this paper, we present some theoretical results on the fixed points of the adaptive iterative soft-thresholding algorithm, with median absolute deviation as the thresholding strategy. 
We prove the existence, non-uniqueness, and stability of the fixed points. 
We also prove the local linear convergence of the algorithm around any stable fixed point and show the global piecewise convergence behavior. 
Our assumptions are minimal throughout the derivation, and we only use deterministic arguments, because we focus on the sparse linear inverse problems with deterministic measurement matrices. 
Through numerical experiments, we verify that ISTA with the MAD can implicitly select appropriate $\lambda$ at various noise levels and is approximately an asymptotic consistent estimator, even for problems with deterministic measurement matrices. The performance also matches the optimal approximate message passing algorithm for a compressive sensing problem.  

Empirically, the ISTA with the MAD seems to be a stable and consistent method for solving sparse linear inverse problems, especially those with deterministic measurement matrices. Many more theoretical results remain to be proved, including but not limited to 
\begin{enumerate}
    \item With a better understanding of the eigenvalues change of nonsymmetric rank-1 update of positive definite matrices, we can prove
    \begin{itemize}
        \item A necessary and sufficient condition for the stable fixed points (recall Proposition \ref{prop_stability} is a necessary condition),
        \item A better global convergence analysis or guarantee.
    \end{itemize}
    \item With more assumptions on measurement matrix $A$ and sparsity level, we can derive a upper bound between two $\lambda$s corresponding to one $\gamma$, and form an almost one-to-one correspondence (per discussion in Remark \ref{one_to_one}).   
     \item Explanation of its consistent performance, especially for the problems with deterministic measurement matrices. This would require understanding the behaviors and properties of the soft-thresholding operator with the median absolute deviation. 
\end{enumerate}

\bibliographystyle{ieeetr}
\bibliography{IEEEabrv,main}

\end{document}